\documentclass{article}

    \PassOptionsToPackage{numbers, compress}{natbib}

\usepackage[preprint]{neurips_data_2021}

\usepackage[utf8]{inputenc} 
\usepackage[T1]{fontenc}    
\usepackage{hyperref}       
\usepackage{url}            
\usepackage{booktabs}       
\usepackage{amsfonts}       
\usepackage{nicefrac}       
\usepackage{microtype}      
\usepackage{xcolor}         
\usepackage{graphicx}
\usepackage{subcaption, upgreek}
\usepackage[normalem]{ulem}
\usepackage{enumitem}
\usepackage{multirow}
\everypar{\looseness=-1}

\title{UltraMNIST Classification: A Benchmark to Train CNNs for Very Large Images}

\author{Deepak K. Gupta$^{\dagger*}$, Udbhav Bamba$^{\dagger}$\thanks{These authors contributed equally.}, Abhishek Thakur$^{\ddagger*}$, Akash Gupta$^{\dagger}$, \\
\textbf{Suraj Sharan$^{\dagger}$, Ertugrul Demir$^{\mathparagraph}$, and Dilip K. Prasad$^{\mathsection}$} \\
$^{\dagger}$Transmute AI Lab (Texmin Hub), Indian Institute of Technology, ISM Dhanbad, India \\
$^{\ddagger}$Hugging Face \\
$^{\mathparagraph}$Global Maksimum Data \& Information Technologies \\
$^{\mathsection}$Department of Computer Science, UiT The Arctic University of Norway, Tromso, Norway}

\begin{document}

\maketitle

\begin{abstract}

Convolutional neural network (CNN) approaches available in the current literature are designed to work primarily with low-resolution images. When applied on very large images, challenges related to GPU memory, smaller receptive field than needed for semantic correspondence and the need to incorporate multi-scale features arise. The resolution of input images can be reduced, however, with significant loss of critical information. Based on the outlined issues, we introduce a novel research problem of training CNN models for very large images, and present ‘UltraMNIST dataset’, a simple yet representative benchmark dataset for this task. UltraMNIST has been designed using the popular MNIST digits with additional levels of complexity added to replicate well the challenges of real-world problems. We present two variants of the problem: ‘UltraMNIST classification’ and ‘Budget-aware UltraMNIST classification’. The standard UltraMNIST classification benchmark is intended to facilitate the development of novel CNN training methods that make the effective use of the best available GPU resources. The budget-aware variant is intended to promote development of methods that work under constrained GPU memory. For the development of competitive solutions, we present several baseline models for the standard benchmark and its budget-aware variant. We study the effect of reducing resolution on the performance and present results for baseline models involving pretrained backbones from among the popular state-of-the-art models. Finally, with the presented benchmark dataset and the baselines, we hope to pave the ground for a new generation of CNN methods suitable for handling large images in an efficient and resource-light manner. 
\end{abstract}

\section{Introduction}
Convolutional neural networks (CNN) are looked at as one of the biggest breakthroughs for image processing, especially because of their capability of extracting information beyond what can be achieved by the conventional computer vision methods (see comprehensive reviews in \cite{khan2020air, Zewen2021tnn, Alzubaidi2021jbd}). However, there are certain limitations of CNN that still need attention, and we address one such aspect in this paper. 


There is an evolution in scientific domains such as microscopy \cite{khater2020review,schermelleh2019super}, earth sciences \cite{huang2018agricultural,amani2020google}, and space observations using the recently launched James Webb space telescope where huge images are created due to the latest scientific technologies. Here, we will use super-resolution microscopy (more popularly referred to as nanoscopy) as a case of discussion, which won Nobel Prize in Chemistry in 2014. Indeed, deep learning is considered of interest in these domains as well, in particular in the microscopy domain \cite{orth2017microscopy,dankovich2021challenges,sekh2020learning,sekh2021physics}. However, the big data challenge of applying CNNs to analyse the nanoscopy images is immense as we demonstrate here.

\begin{figure}
\centering
\includegraphics[scale=0.265]{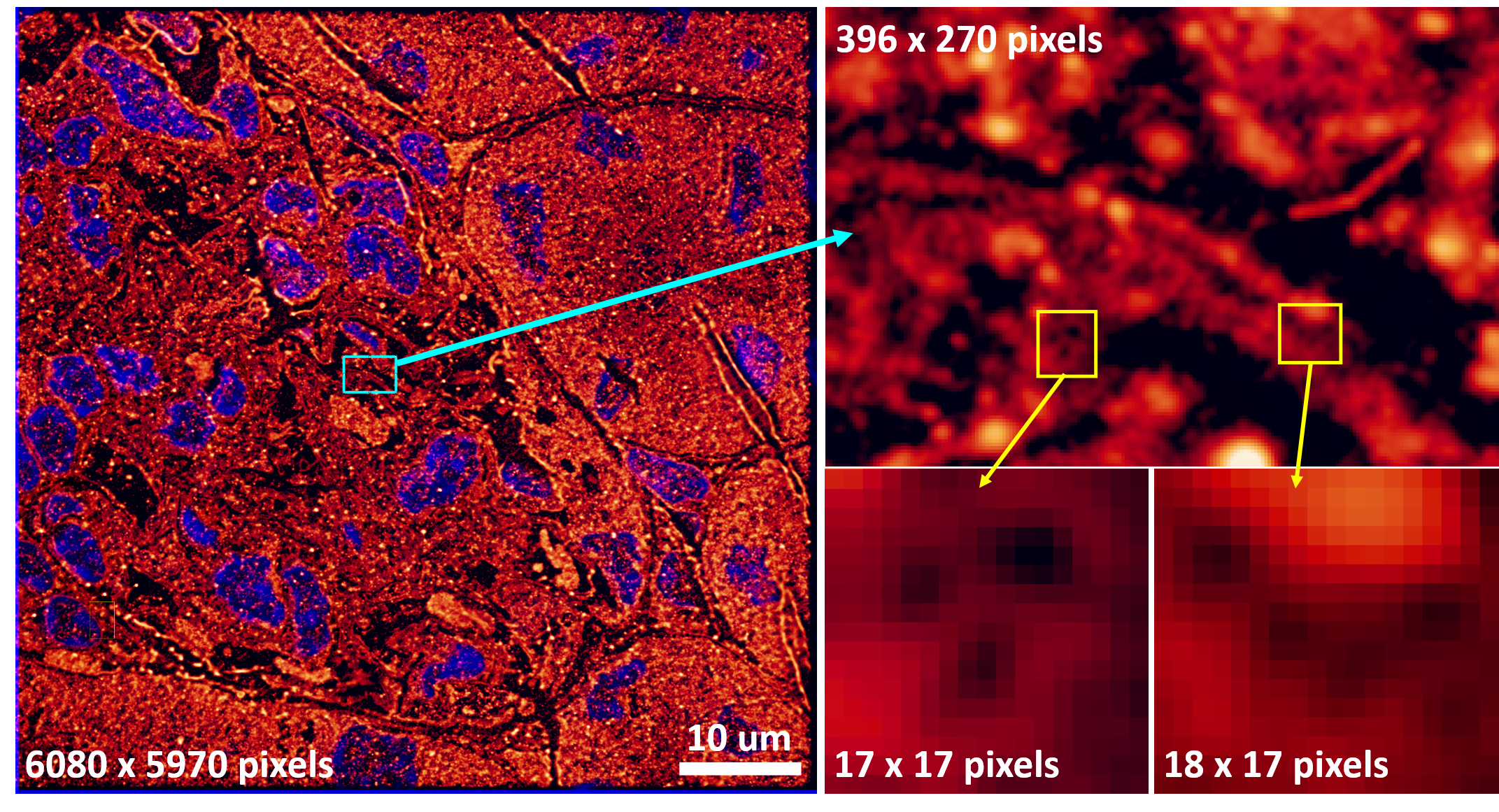}  
\caption{On the left is a nanoscopy image of mouse kidney cryo-section approximately 1/12th of the area of a single field-of-view of the microscope, chosen to illustrate the level of details at different scales. The bottom right images show that the smallest features in the image of relevance to biological factors can be as small as a few pixels (here 5-8 pixels for the holes)\cite{villegas2022chip}.} 
    \label{fig:STORM}
\end{figure}



High content nanoscopy involves taking nanoscopy images of several adjacent field-of-views and stitching them side-by-side to have a full perspective of the biological sample, such as a patient's tissue biopsy, put under the microscope. There is information at multiple scales embedded in these microscopy images \cite{villegas2022chip}, such as shown in Fig. \ref{fig:STORM}, with the smallest scale of features being only a few pixels in size. Indeed such dimensions of images and such levels of details is a challenge for CNNs which are often conventioanally designed for a fixed input size, such as 128$\times$128 pixels or up to 512$\times$512 pixels. 


With these large images, there are only very limited ways to approach the problem using current CNN methods. It is most common to reduce the resolution of the image, however, the small scale features are lost and this can completely change the semantic context associated with the images. Another popular strategy is to break the image into a set of partially overlapping tiles and train CNN with subsets of tiles. The limitation is that the semantic link between the tiles is not known \emph{a priori} and disconnecting them leads to loss of semantic information across the image. This issue can be resolved to a certain extent using a 2-stage training pipeline where the original image is first projected onto a low-dimensional latent space and then this compressed version is used for training. However, this approach as well involves significant information loss and can only work for up to a certain level of increased size of the input. With the extreme variation in scale of the features, choosing the right tile size is also hard. Multiple CNNs could be used to handle different scales, or methods need to be developed to collate information from the outputs of all these CNNs into a single coherent output. The challenges in 3D and 3D with temporal data are further compounding in terms of the handling of data, designing the right architectures, and incorporating spatial and temporal correlations of value.

\begin{figure}
    \centering
    \begin{subfigure}{0.22\linewidth}
    \fbox{\includegraphics[scale=0.018]{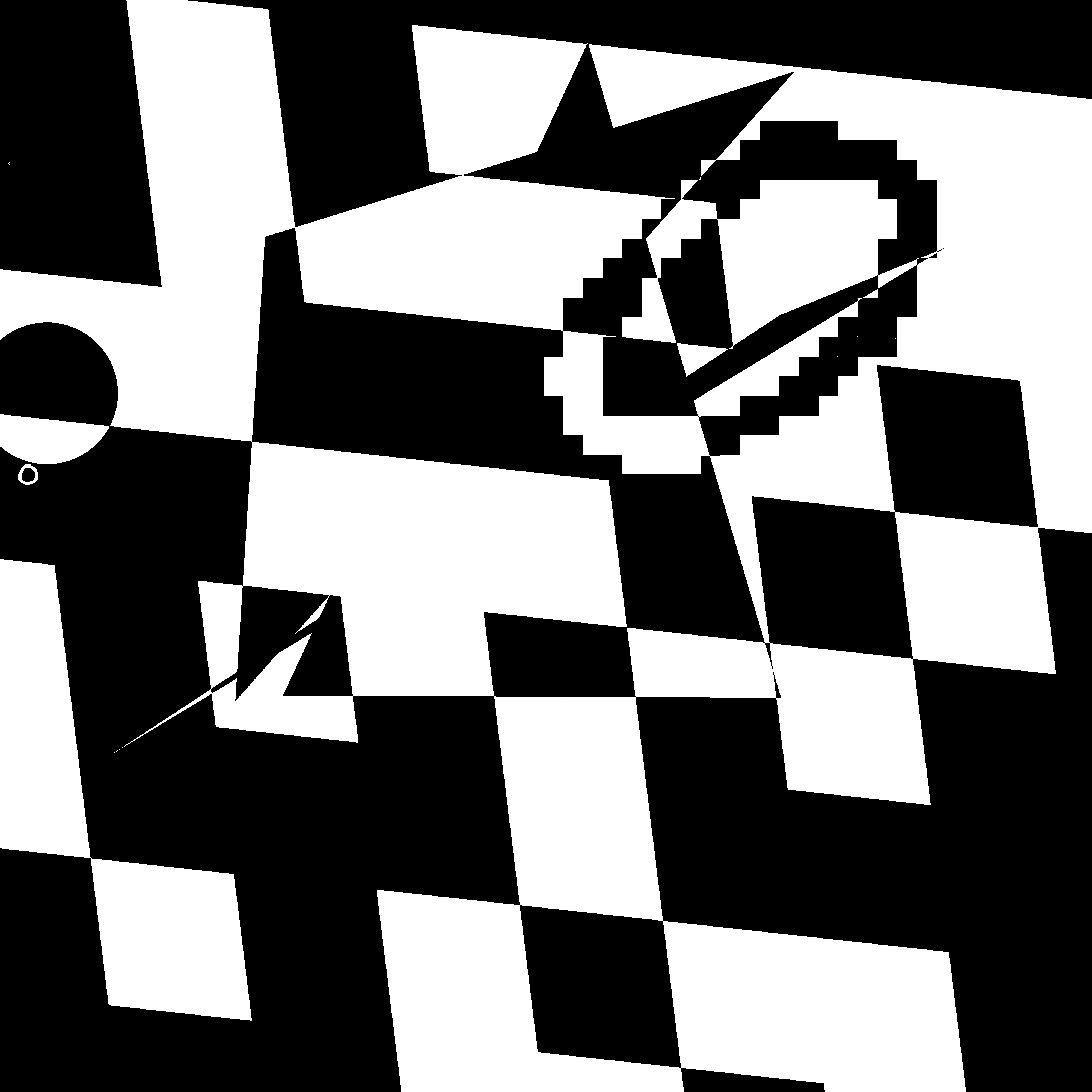}}
    \caption{Class: 0}
    \end{subfigure}
    \begin{subfigure}{0.22\linewidth}
    \fbox{\includegraphics[scale=0.018]{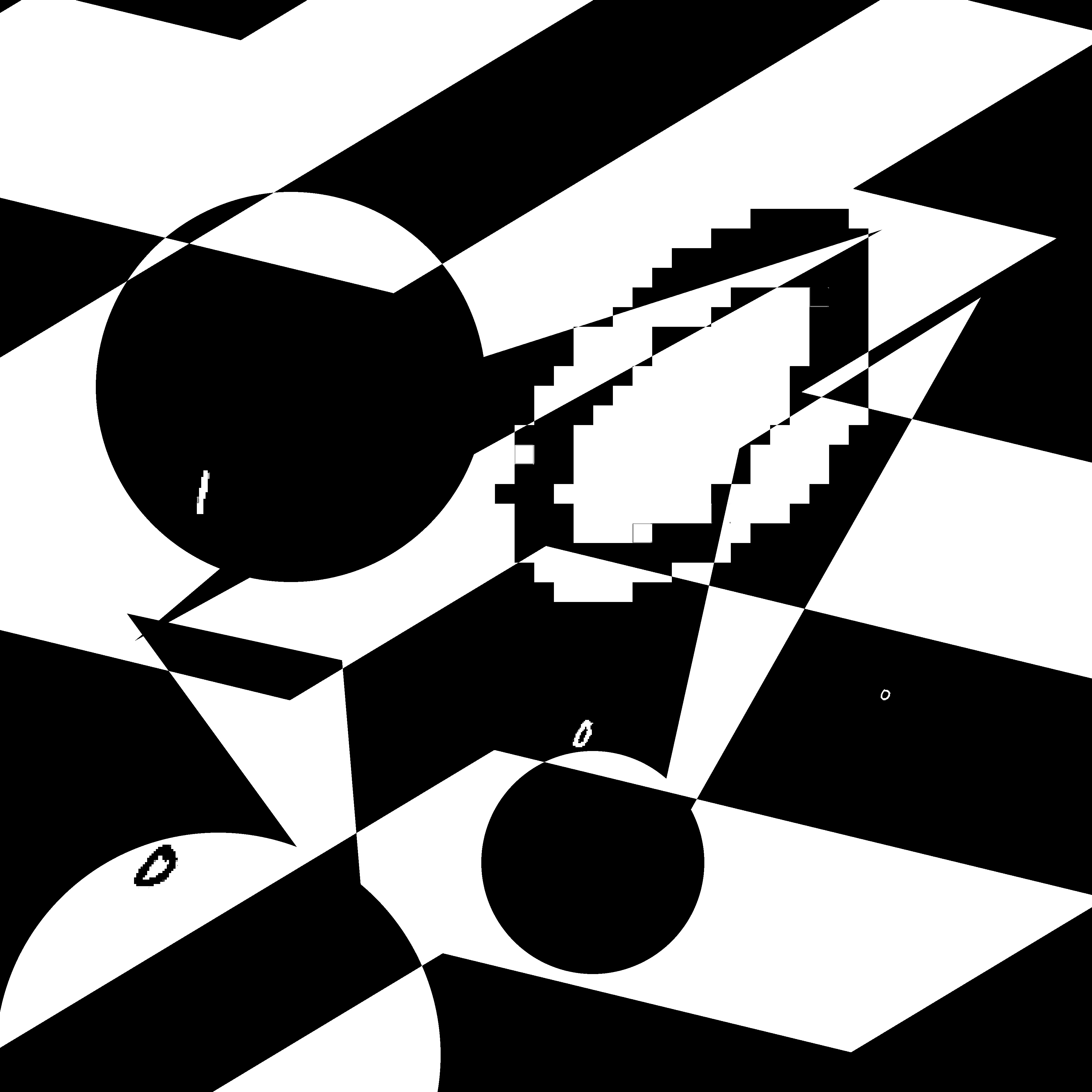}}
    \caption{Class: 1}
    \end{subfigure}
    \begin{subfigure}{0.22\linewidth}
    \fbox{\includegraphics[scale=0.018]{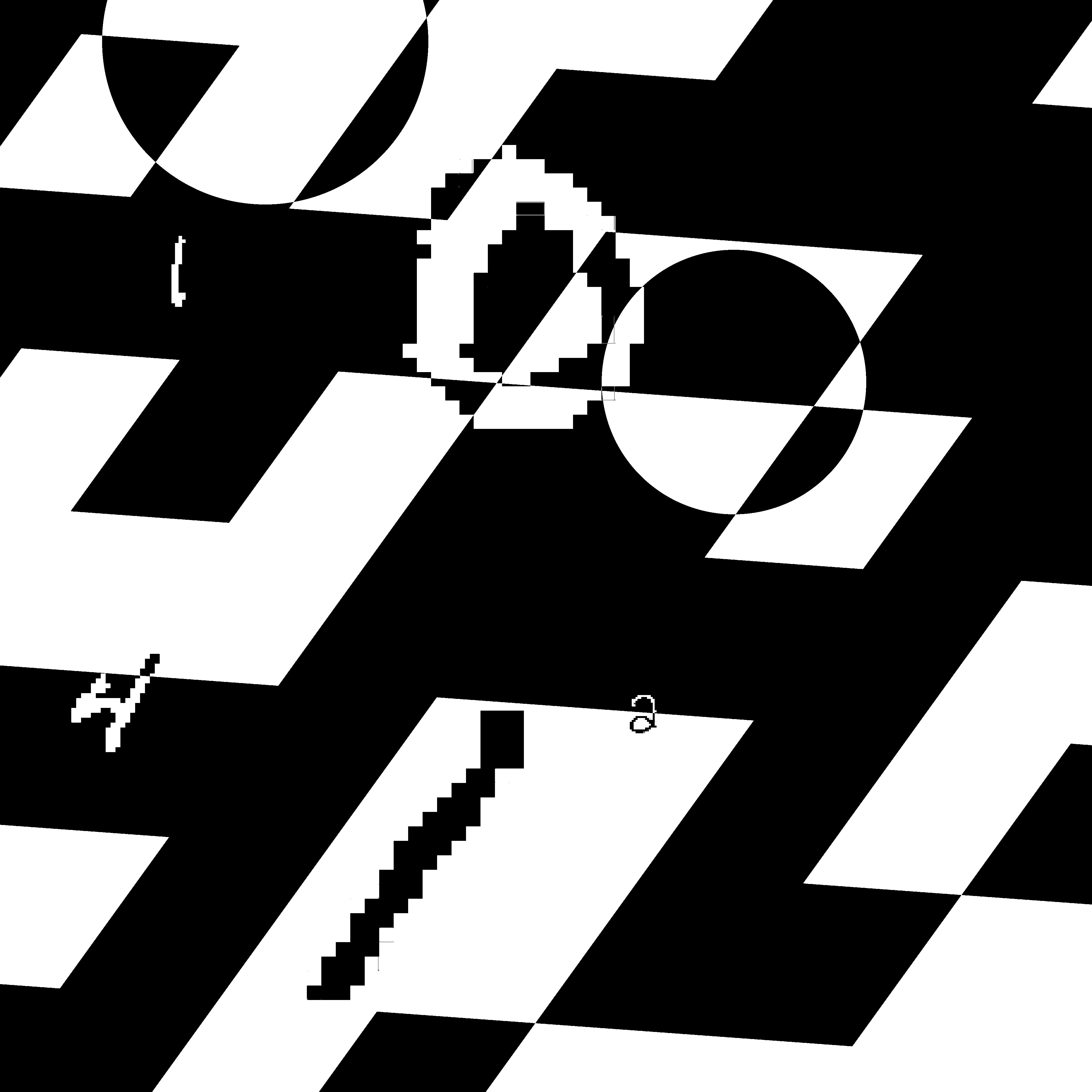}}
    \caption{Class: 8}
    \end{subfigure}
    \begin{subfigure}{0.22\linewidth}
    \fbox{\includegraphics[scale=0.018]{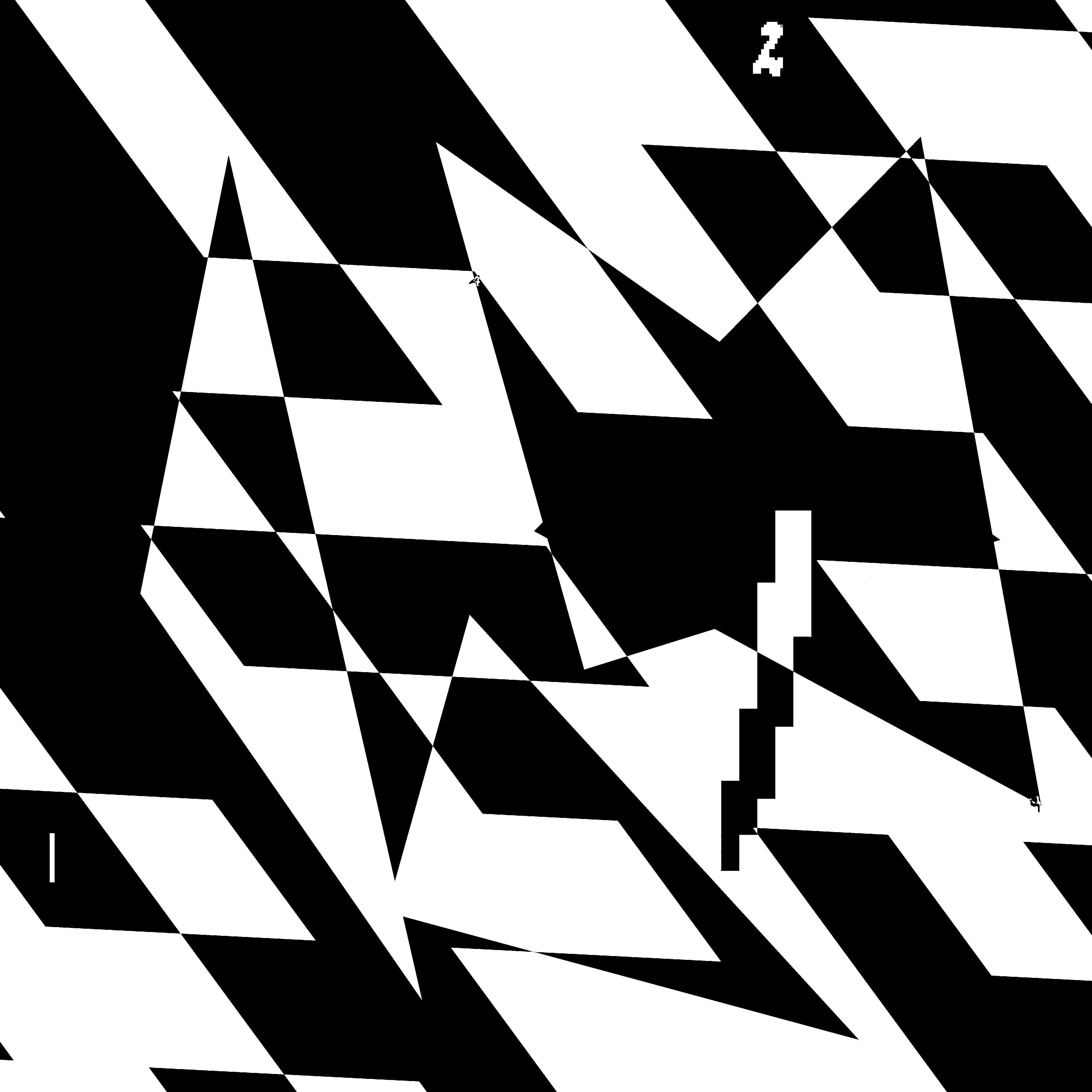}}
    \caption{Class: 12}
    \end{subfigure}\\
    \begin{subfigure}{0.22\linewidth}
    \fbox{\includegraphics[scale=0.018]{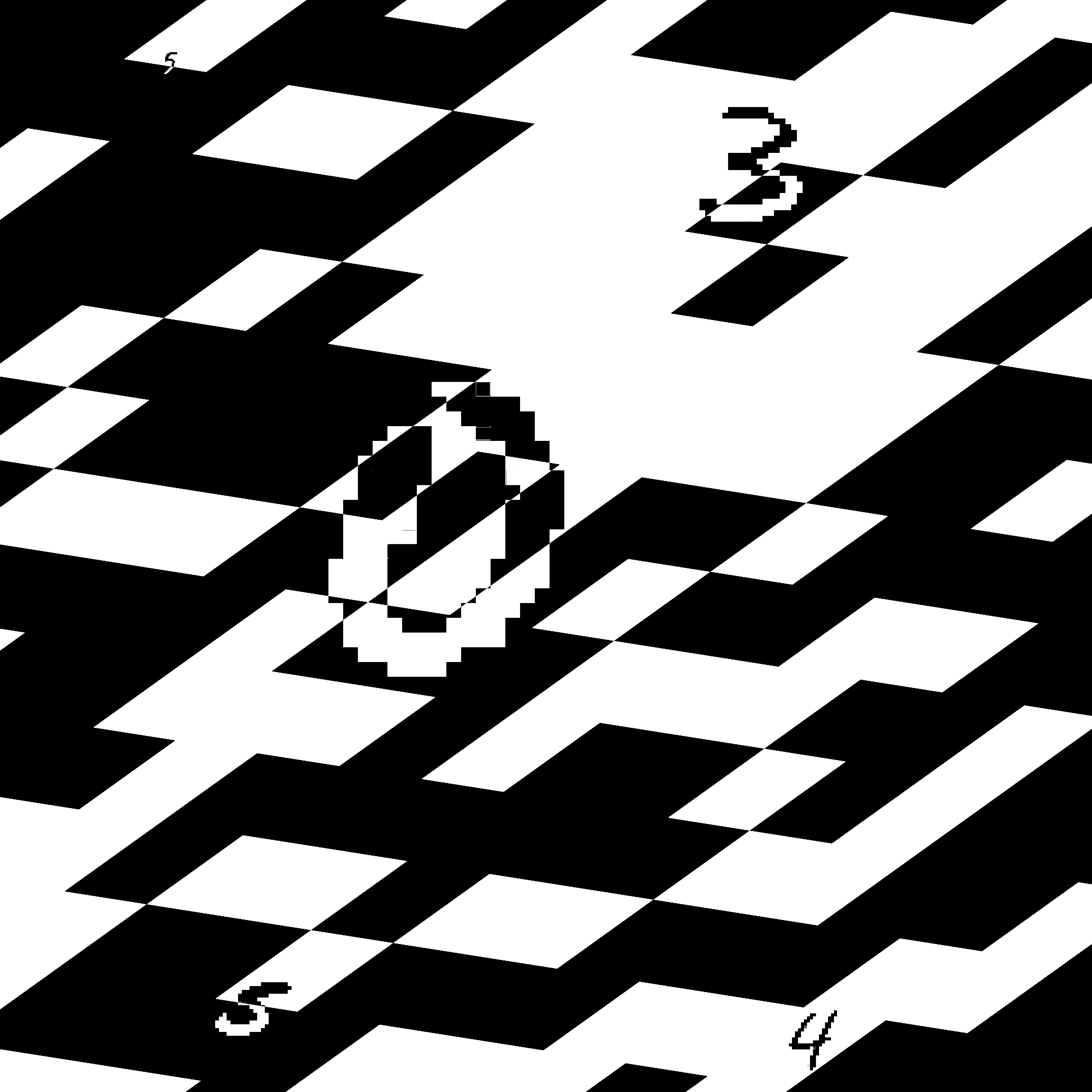}}
    \caption{Class: 17}
    \end{subfigure}
    \begin{subfigure}{0.22\linewidth}
    \fbox{\includegraphics[scale=0.018]{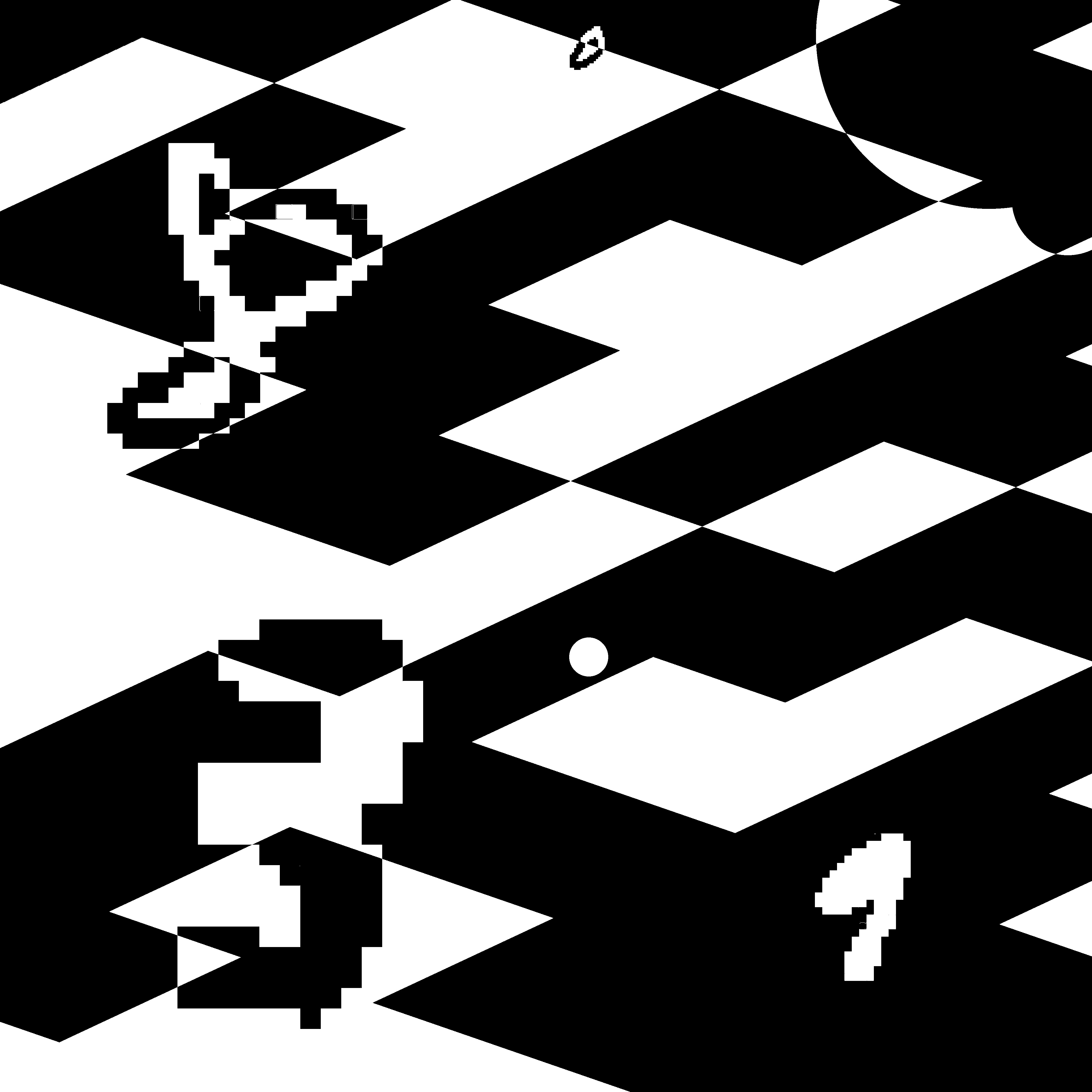}}
    \caption{Class: 20}
    \end{subfigure}
    \begin{subfigure}{0.22\linewidth}
    \fbox{\includegraphics[scale=0.018]{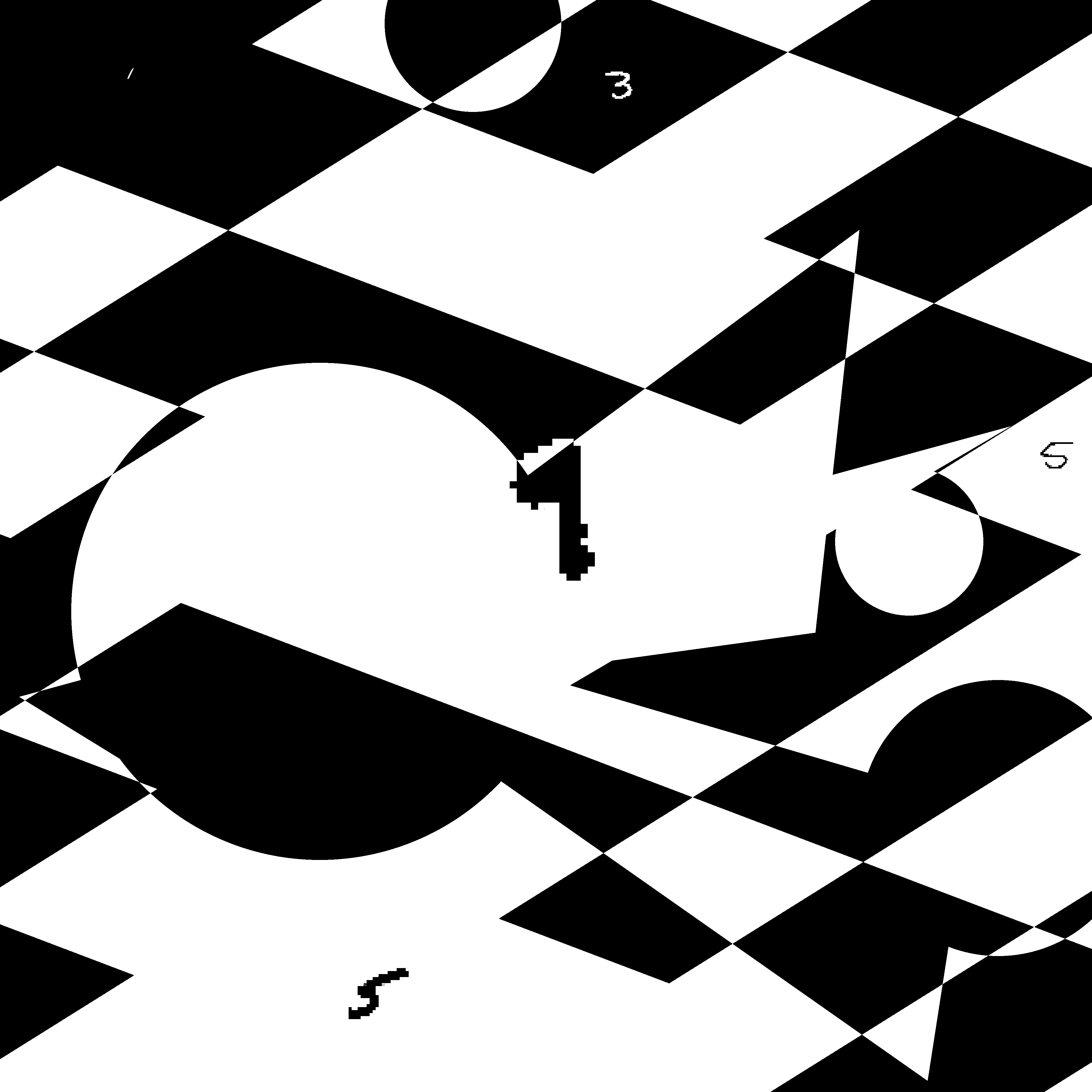}}
    \caption{Class: 23}
    \end{subfigure}
    \begin{subfigure}{0.22\linewidth}
    \fbox{\includegraphics[scale=0.018]{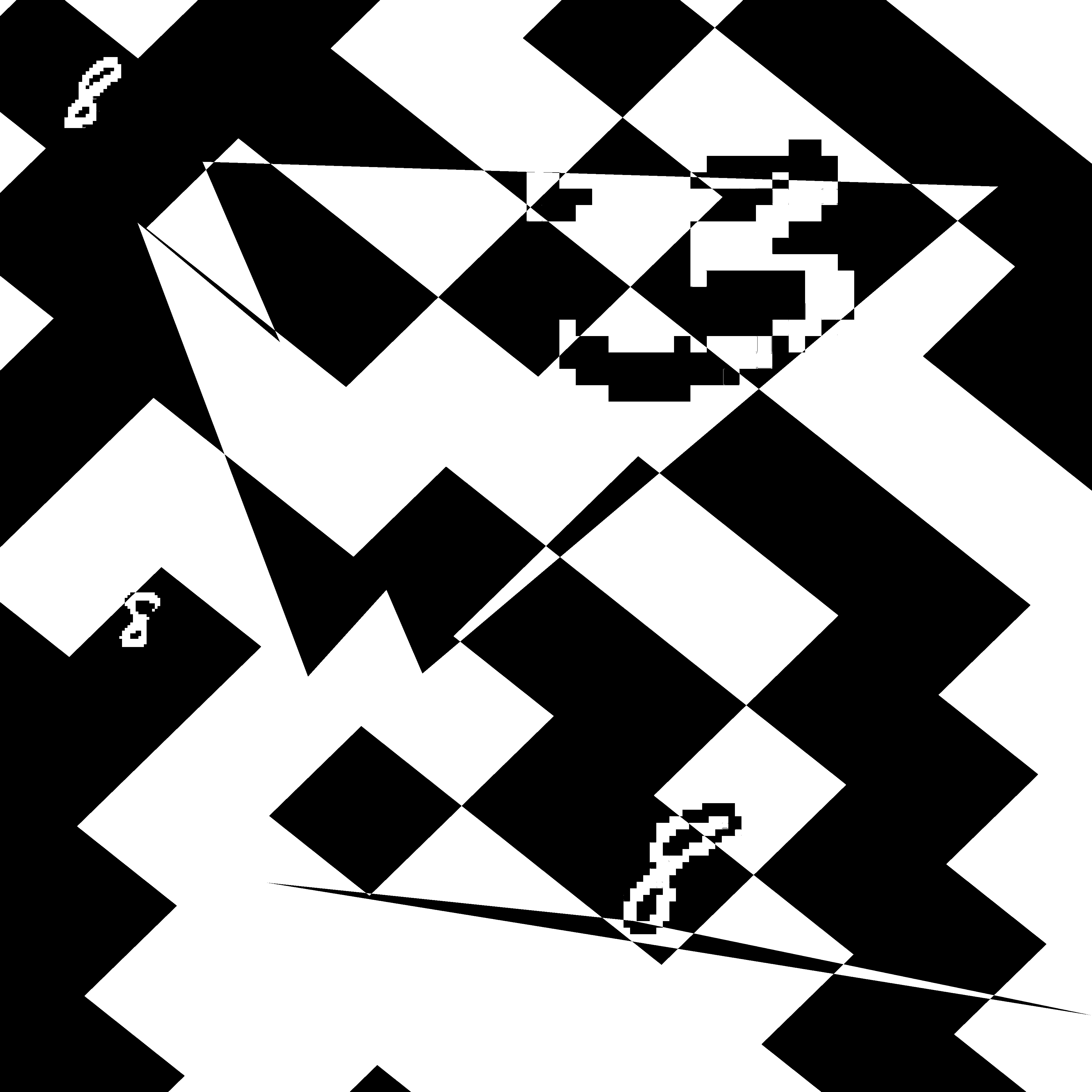}}
    \caption{Class: 27}
    \end{subfigure}
    \caption{Example images for 8 out of the 28 classes of UltraMNIST dataset.}
    \label{fig-umnist-samples}
\end{figure}

To promote the development of solutions to the problem outlined above, we introduce `UltraMNIST dataset', a simple get good representative dataset to develop CNN training methods for handling very large images. UltraMNIST data samples are constructed using the popular MNIST digits \cite{LeCun2005TheMD} with some additional modalities. Each sample image is a $4000 \times 4000$ grid and contains 3-5 MNIST  digits varying in size from as small as $14 \times 14$ pixels to as large as $2000 \times 2000$ pixels.  Example sample from this dataset are shown in Fig. \ref{fig-umnist-samples}. 
Further, a complex stochastically sampled background is added to restrict explicit sampling of the MNIST digits from the full scale UltraMNIST sample. Details related to the synthesis of the background are described in a later section of this paper. The machine learning problem to be solved here is to predict the sum of all the digits that exist in every image. This poses the problem of building a semantic relation between distant parts of the images such that the features in different parts are significantly different in terms of scale.

We present two variations of the UltraMNIST benchmark. First is the standard version where there is no restriction on the GPU or TPU memory to be employed for the problem. This is intended to promote the efficient use of the latest hardwares and identify the best CNN training strategies that could even involve splitting the training of a large image on a distributed set of GPUs. Alternatively, it could also be possible to develop smart training or downsampling strategies that deliver good performance with even the simple models such as ResNet-50, and the standard UltraMNIST classification benchmark provides this opportunity. On a different note, UltraMNIST digits are representative of even larger images which could be even as large as $50000 \times 50000$ pixels and more, and even the state-of-the-art GPU resources cannot work with this resolution. To mimic this challenge of the real problems, we introduce a `budget-aware variant' of the UltraMNIST classification problem. This benchmark puts a constraint on the GPU memory that can be used by the training and inference steps of any novel solution to the UltraMNIST classification problem. In this paper, we choose a memory budget of 11 GB to build the first baselines.

To accelerate the development of useful models, we present several baseline solutions for the two variants. These methods explore the how well the current SOTA classification models work for the presented problem. To keep the baselines competitive, we hosted a Kaggle community competition\footnote{Hosted at \texttt{https://www.kaggle.com/c/ultra-mnist}} where we invited the community of data scientists to propose novel solutions to UltraMNIST classification. Most of the solutions are based along the lines of reducing the resolution of the large images and seeing the extent of drop observed with different SOTA models. To understand the efficacy of the machine learning models, we also present a human baseline score computed based on labelling of a subset of the test data by humans. Further, solutions are presented for different resolutions of the input for popular models such as ResNets and EfficientNets. 

\textbf{Contributions.}
\begin{itemize}[noitemsep]
\item In this paper, we introduce a novel research problem of training CNN models for very large images. Referred as `UltraMNIST classification', this problem is designed to innovate in terms of novel CNN training pipelines that can facilitate the efficient handling of very large images under limited GPU memory constraints.
\item We present `UltraMNIST dataset', a simple yet good representative benchmark dataset for the problem outlined above. UltraMNIST has been designed using the popular MNIST digits with additional levels of complexity added to replicate well the challenges to be addressed in the real-world problems. 
\item We present two variants of the classification problem: `UltraMNIST classification' and `Budget-aware UltraMNIST classification'. The standard ultramnist classification benchmark is intended to facilitate the development of novel CNN training methods that make the effective use of the best available GPU resources. The budget-aware variant is intended to promote development of methods that work under constrained GPU memory.
\item For the development of competitive solutions to the UltraMNIST classification problem, we present several baseline models for the standard benchmark and its budget-aware variant.
\item We study the effect of reducing resolution on the performance of the model, and results are presented for baseline models involving pretrained backbones from among the popular state-of-the-art models.
\end{itemize}

\textbf{Implications. } We hope that the research work presented in this paper promotes further development on the following aspects.

\textit{Enhanced interpretation of very large images. }With the presented benchmark dataset and the associated baselines, we hope to see the development of novel training strategies for CNNs that do not involve any downsampling of the input images. Solutions to such problems are of significant value to domains such as medical imaging and aerial imagery analysis, and we hope that apt solutions to the presented problem could add value in terms of extracting better semantic information as well as processing under limited computational memory resources when input images are very large.

\textit{Building extreme scale-agnostic models. }Solutions to the UltaMNIST problem could answer the question of handling extreme scale variations in any image. While CNNs learn invariance/equivariance to a certain extent, doing the same for scale variations of beyond 1:1000 or more is currently not possible, and we hope the solutions to UltraMNIST could help achieve this.

\textit{Translating computational memory load into computational time load. }In the current state, processing of very large images with CNNs leads to a bottleneck in terms of computational memory usage. Hardwares have fixed amount of GPU memory, and CNN models processing very large images cannot fit in this memory space. We speculate that UltraMNIST benchmark, and especially its budget-aware version, will lead to the development of new CNN training methods that scale the computational load in time dimension such that even very large problems can be handled with limited GPU memory.

\section{Related Work}

We discuss here some of the research works that come close to the UltraMNIST dataset and also clearly state how our benchmark differs from them. 

There exist several high-resolution datasets for the task of object detection or segmentation. Some such examples of object detection datasets include High-Resolution Salient Object Detection (HRSOD) dataset \cite{Zeng2019TowardsHS}, NOD \cite{Morawski2021NODTA} dataset, and DOTA dataset \cite{9560031}. Among these, DOTA comprises images of up to resolution as high as 20000$\times$20000 with objects of varying scales, orientations and shapes. Although all these datasets comprise high-resolution images, these do not exhibit any semantic relation between the different objects, and can be easily tackled using approaches such as tiling and aggregating, among others. Among the high-resolution datasets for segmentation are CamVid \cite{BrostowSFC:ECCV08, BrostowFC:PRL2008}, SYNTHIA \cite{Ros_2016_CVPR} and  High-Resolution Fundus (HRF) dataset. The resolution of first two datasets is still not very high, and conventional CNN-based methods can be adapted. HRF dataset comprises images of  3304$\times$2336 size, however, the semantic relation between objects-of-interest is still very local. Thus, HRF as well can be handled using tiling and aggregating methods. 

There exist high-resolution datasets for other computer vision tasks as well. For example,  Flickr-Faces-HQ (FFHQ) \cite{FFHQ} consists of high-quality PNG images at 1024$\times$1024 resolution and is used for GAN-based methods to generate faces. Another is the dataset on single-image super-resolution from NTIRE 2017 challenge \cite{Timofte_2017_CVPR_Workshops}. It comprises images of 2K resolution and is used for the task of super-resolution. However, both these datasets have a completely different objective since even their low-resolution variants would hardly lose any important semantic context.  

Two variants of the MNIST dataset close to our UltraMNIST benchmark are cluttered MNIST and large-scale MNIST \cite{jansson_ylva_2020_3820247}. Cluttered MNIST comprises MNIST digits with added clutter on a low-resolution grid. Large-scale MNIST comprises images with large scale variation in the MNIST digit size. However, neither there exists any strong semantic correspondence in different parts of these images, nor the resolution is too high. Thus, conventional CNNs with small adaptions can already work well on these datasets.

In terms of the similarity, the closes to our dataset is PANDA dataset, comprising very high-resolution images of prostate tissue samples, and the goal is to estimate severity of the disease. This is a classification problem that requires learning strong semantic relation between different parts of the large images. However, the dataset is very huge and the labels are imperfect. UltraMNIST draws inspiration from this large-scale dataset. Through the use of MNIST digits and posing it as a simpler classification problem, UltraMNIST benchmark provides a platform for developing CNN training methods which in turn could be improved further for the PANDA dataset as well.

In terms of handling large images, there are two specific categories, the first being \textit{hardware-specific}, whose application ranges from data and model parallelism \citep{Le2019} to hardware processing. A typical solution applies \textit{divide and conquer principle} as it splits the images in tiles and process each of them individually using CNNs and put these tiles back together using normalization and padding techniques \citep{Wu2017acm}. Although, such approaches can be consume less memory as each patch is processed separately, they also give sub-optimal performance due to information loss. The second approach is \textit{Architecture Specific} which attempts to decode feature space of the distribution and at the same time use image compression algorithms paired with some post processing techniques \citep{Wu2017acm, Talebi_2021_ICCV} for generalization. These methods as well tend to lose the information related to very fine details in the high-resolution images. We hope that with the UltraMNIST benchmark, there will be developments towards novel CNN training pipelines that train even the very large images in a end-to-end manner without the need for downsampling.

\section{UltraMNIST Classification Benchmark}

\label{sec:ultrabenchmark}
\subsection{UltraMNIST dataset}
\begin{figure}
    \centering
    \includegraphics[scale=0.52]{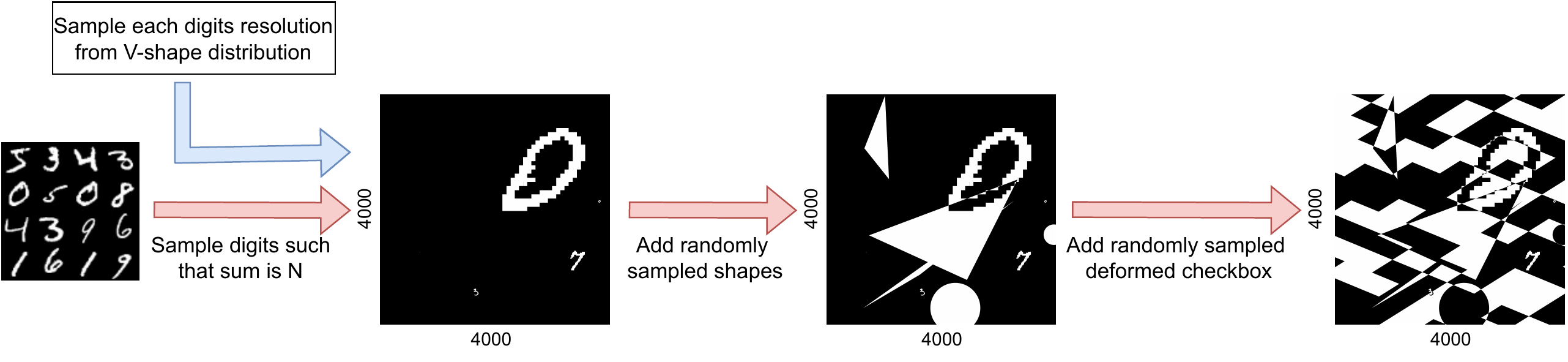}
    \caption{Schematic workflow for the generation of UltraMNIST samples.}
    \label{fig-umnist-dataset}
\end{figure}

\textbf{Data description. }UltraMNIST comprises a total of 28000 images in the training set as well as 28000 images in the test set. Each UltraMNIST sample comprises 3 to 5 MNIST digits, and the sum of the digits forms the label of the sample. The digits are sampled such that the minimum sum is 0 and the maximum sum is 27. Beyond the MNIST digits, UltraMNIST samples also comprise random shapes and a randomly oriented checkerboard pattern for additional complexity. The goal is to build a classifier model that classifies each UltraMNIST sample into one of the 28 classes based on the sum of the digits contained in it.

\textit{Why MNIST digits? }The primary reason to use MNIST as the base dataset is that the digits of this dataset are relatively simple to classify and require only a few discriminative features to be represented well. With the additional complexities, the MNIST digits clearly suffice the requirement of complexity needed for our task. With this, we reduce the load of building a too complex and heavy discriminative model to identify the objects in our UltraMNIST samples, so that more effort can be put on how the CNN training pipeline itself can be improved for handling our large images.

\textit{Why 3-5 digits per UltraMNIST image? }For semantically connecting different parts of the image, we needed at least two samples. To add to the difficulty level, we wanted to choose a higher number, however, going beyond 5 added too much clutter, hence a range of 3 to 5 was chosen.

\textit{Why extreme scale contrasts? }The motivation for using extreme scale contrasts of as much as 1:140 comes from the requirements posed by real-world problems. Additionally, the scale contrast prohibits reducing the resolution of the images to a very low order since the small digits would disappear then and the semantic context would be lost. The too large digits are chosen such that the receptive field of the kernels employed in current CNN training schemes cannot capture them fully.

\textit{Why background clutter? }If the digits are left on an empty white background, they can be individually extracted using simple pre-processing methods and brought to the same scale. While this makes the problem easy to solve, it will completely ruin the whole motivation of this research problem. To circumvent this issue, we have added a complex background which cannot be simulated easily, and ensure that any preprocessing object detection methods are not used.

\textbf{Creation method. }The process of creation of UltraMNIST samples is schematically described in Fig. \ref{fig-umnist-dataset}. First, 3 to 5 digits are randomly sampled from the MNIST training set. Next, for each MNIST, a scale value is sampled from a V-shape distribution, where the two extremes are $14 \times 14$ and $2128 \times 2128$. Using these scales, the sampled digits are transformed and placed without overlap on a blank canvas of $4000\times 4000$ resolution image. 

To further increase the difficulty of UltraMNIST and to make it ideal for real-world benchmarking, we add a complex background in each image. The addition of background is a two fold process where firstly we sample random circles and triangles and overlay randomly them using XOR operation on the canvas. We sample each shape randomly from zero to five. Further,  an augmented checkerboard pattern is sampled which is then overlaid on the canvas. Augmentations on the checkerboard include rotations by $\pm$45\% and shear of $\pm$50\%.

\subsection{UltraMNIST benchmark}
UltraMNIST benchmark is a classification problem that aims at assigning each UltraMNIST digit to one of the 28 classes defined earlier. The foundation of UltraMNIST classification benchmark is built around the popular MNIST classification problem with added levels of complexity to pose the challenges of very large images carrying distributed semantic context. The goal is to develop novel CNN training strategies to achieve highest classification accuracy score on the test subset of UltraMNIST. This benchmark promotes two different directions of research: First is to modify the current CNN training pipelines such that the current classification models can be used for the classification of very large images. Second is to modify the architecture of current CNN models to adapt their receptive fields for very large images and deploy them on the most recent state-of-the-art GPU hardwares. Scientific developments along both the lines answer the question posed by this benchmark and are to be treated as acceptable solutions. Based on the description above, we formally describe this benchmark problem as follows.

\textit{\underline{UltraMNIST Classification Benchmark}. Find a solution to the classification of UltraMNIST digits into one of the target classes such that the accuracy score obtained over the test dataset is maximized.}

Note that the use of the MNIST dataset in any form during training or testing of UltraMNIST digits is prohibited. The primary reason is that the UltraMNIST digits are designed to mimic real-world problems where a single object in a large image might not necessarily carry any semantic context, but collectively multiple objects in that image make a meaning. We pose a similar problem where no a priori mathematical information on the MNIST digits is to be hard baked into the model. The trained models should not be explicitly made to understand the MNIST digits, rather only the labels of the UltraMNIST digits should be only be used to guide the supervised learning process.  For example, a single digit `5' in our images should not be interpreted as `5' since there is no information as such in the data. However, occurrence of sets such as `5, 0, 0', `4, 1, 0' and `3, 1, 1' are to be labelled as `5'. 

On a different note, we also accepted solutions to the unconstrained version of this problem in the challenge hosted on the Kaggle platform, and there were several interesting solutions proposed that used the original MNIST digits. For an overview of the unconstrained ultramnist classification variant, see Appendix \ref{sec-app-results} of this paper.

\subsection{Budget-aware UltraMNIST Benchmark }

Budget-aware UltraMNIST classification refers to a similar classification problem as described above, but with additional constraint on the usage of GPU memory during training and inference of the deep learning model. With the additional constraint, we hope that novel solutions will be proposed that develop CNN training pipelines specific to available hardware specifications. In general, hardware-related developments are slower than the software ones, and it is not an easy task to upgrade the GPU hardware based on the requirements posed by changes in CNN models. We propose to investigate methods that alleviate this challenge through transforming the load on computational memory into increased computational time. This would hopefully allow to train CNN models for very large images with even constrained GPU resources. An added advantage would be that the novel strategies developed using this benchmark could help to even scale the current large CNN models for low memory devices. In this paper, we present the case of GPU memory budget of 11 GB. However, the concept of budget-aware classification of UltraMNIST holds for any prescribed budget, and we present scores for the GPU memory budget of 24 GB in Table \ref{table-budumnist} in the appendix section of this paper. Further, we express this problem more formally as follows.

\textit{\underline{Budget-aware UltraMNIST Classification Benchmark}. With the maximum GPU memory usage of N GB during training and inference stages, find a solution to the classification of UltraMNIST digits into one of the target classes such that the accuracy score obtained over the test dataset is maximized.}

\section{Baselines}

\subsection{Human baseline}

UltraMNIST poses the challenge of identifying very small MNIST digits in the midst of complex background noise. Unlike the original MNIST digits, our UltraMNIST is also hard for humans to label. To validate this, we first present our human baseline developed for this dataset. For this, we approached several people who were not involved in the curation process. Each individual was shown the entire set of MNIST digits that were used for curation, and asked to only search for these in the images and sum them up to form the corresponding label. Further, we explained them on the contrasting scale of the digits from being extremely small to very large. Lastly, we also informed them that each image could contain 3-5 MNIST digits. Based on all this information, candidates were asked to label between 20-40 random images from the test set. In total, we got a subset of 1000 images labelled and an accuracy score of 52.1\% was obtained (see Table \ref{table-umnist}).

\begin{table}[]
    \centering
    \begin{tabular}{l | l | c | c | c}
         \textbf{Method} & \textbf{Backbone} & \textbf{Input size} & \textbf{Acc. (\%)} & \textbf{Acc.@11 GB (\%)}\\
         \toprule
         Random baseline & - & - & 3.6 & -\\
         Human baseline & - & - & 52.1 & - \\
         \midrule 
         RN-256 & ResNet-50 & 256$\times$256 & 8.9 & 8.9\\
         RN-512 & ResNet-50 & 512$\times$512 & 24.5 & 24.5\\
         RN-1024 & ResNet-50 & 1024$\times$1024 & 31.3 & 31.3\\
         ENB0-256 & EfficientNet-B0 & 256$\times$256 & 20.4 & 18.6 \\
         ENB0-512 & EfficientNet-B0 & 512$\times$512 & 49.8 & 43.9 \\
         ENB0-1024 & EfficientNet-B0 & 1024$\times$1024 & 49.0 & 49.0 \\
         ENB3-256 & EfficientNet-B3 & 256$\times$256 & 30.3 & 26.7 \\
         ENB3-512 & EfficientNet-B3 & 512$\times$512 & 72.3 & \textbf{69.1} \\
         ENB3-1024 & EfficientNet-B3 & 1024$\times$1024 & 79.1 & 54.5 \\
         MBNet-demir & MobileNet-V3 & 1280$\times$1280 & 86.2 & -\\
         EffNetTPU-vecxoz & EfficientNet-B7 & 1536$\times$1536& \textbf{93.8} & - \\
         \bottomrule
    \end{tabular}
    \vspace{0.5em}
    \caption{Performance scores for the various baselines on UltraMNIST and Budget-aware UltraMNIST benchmarks. For memory budget, we choose an upper limit of 11 GB GPU memory for the model training and inference tasks.}
    \label{table-umnist}
\end{table}

\subsection{Deep CNN models}

We also present here a set of baselines built using some of the state-of-the-art models used in image classification. To keep these baselines competitive, we hosted a community competition on the Kaggle platform where data scientists from across the globe could participate and submit solutions to the UltraMNIST classification problem. Among the solutions that satisfied the criterion outlined for the UltraMNIST benchmark, we either directly used them or added some fine-tuning to improve them further. Further, we also adapted some of the solutions to satisfy the budget-constraint described in the Budge-aware UltraMNIST benchmark. All scores related to the benchmarks are reported in Tables \ref{table-umnist}. We describe below the details related to each of these methods. 

\textbf{ResNet and EfficientNet variants at several resolutions. }As first CNN baselines, we present models using the popular image classification backbones, namely Resnet-50 \cite{He2016DeepRL}, EfficientNet-B0 \cite{Tan2019EfficientNetRM} and EfficientNet-B3. Three variants of image resolution are tested: $256 \times 256$, $512\times 512$ and $1024\times 1024$. We use the method naming convention in the format `BACKBONE-IMAGE WIDTH', thereby RN-256 denotes an image classifier trained on images of resolution $256 \times 256$. See Table \ref{table-umnist} for the full set of methods. Datasets at different resolutions have been created by resizing the original $4000 \times 4000$ images using inter-area interpolation. To avoid any excessive loss of information, images of $256\times 256$ were obtained in a 2-step downscaling process. First the original images were converted to $512 \times 512$, and then these were used to get the images of resolution $256\times 256$. For all the methods, The output of the backbone is mapped to predefined 28 classes using two fully-connected layers with ReLU activation function between them. See the appendices of this paper for more details.

\textbf{MBNet-demir. }This method is based on the approach presented in \citet{Talebi_2021_ICCV}, where a resizing network is used instead of the conventional interpolation algorithms such as bilinear and bicubic. This resizing network is trained together with the backbone (Mobilenet V3)  network and produces a more "CNN friendly" representation of the image which is in turn fed to the backbone as input. It helps further to preserve information of smaller objects which the traditional approaches misses. This resizing network consist of bilinear resizers, which directly skips the network and adds the resized image to the output of the other branch. The other branch uses 2d convolutions, as bottleneck and user defined number of typical residual blocks, these skip connections provide an easier learning process.

\textbf{EffNetTPU-vexcoz. }This method involves training building a classification model using the EfficientNet-B7 backbone and fine-tuning it on the UltraMNIST training set. It involves two stages of training. During the first stage, the model is trained with images of resolution $1024\times1024$, and in the second stage, training is performed at image resolution of $1536\times1536$. For the two stages, learning rates set to $5 \times 10^{-4}$ and $10^{-4}$ with a reduction on plateau. For the purpose of augmentation, image inversion has been used - new images were created through subtracting from 255. 

\section{Discussion}
\label{sec:discussion}
Based on the results reported for the baselines in Table \ref{table-umnist}, we point out here some interesting insights. As anticipated, reducing the model resolution too much has a large adverse effect on its performance. This is clear from the results reported for the image size of $256 \times 256$ pixels for the various methods. Even with the use of EfficientNet-B3 backbone, the maximum accuracy score achieved is only 30\%. For ResNet-50, it goes to lower than 10\%, clearly implying significant loss of information at this reduced resolution. It is also clear that higher resolution images help to achieve better scores, however, as depicted in Table \ref{table-umnist}, these require more powerful computational resources. In a constrained GPU budget setting, only very small batch size is possible for the training sets of such cases, and this leads to a dip in performance. This can be seen from the scores of ENB3-512 and ENB3-1024, where ENB3-512 outperforms the latter. 

We further see that the EffNetTPU-vecxoz method significantly outperforms the other baselines. The primary reason for this success is the utilization of more than one image resolutions in the training pipeline. Moreover, this method uses very high-resolution images in the second stage and utilizes TPU resources to fine-tune EfficientNet-B7. Nevertheless, such scalability of resources can happen only to a limited extent, and this clearly is not the final solution that we aim for the UltraMNIST benchmark, and especially its budget-aware variant. On the positive side, the take away from this baseline is that in case we choose to work with reduced image resolutions, a direction would be to use more than one image resolutions and build a unified pipeline. As a future research direction, this could also be something to investigate further.

While we have analyzed the performance of several baseline methods as well as their combination with different input sizes, it is also of interest to see the challenges posed by the semantic content of the UltraMNIST samples. Due to space constraint, we present this study in the supplementary section of this paper. We look at the results of ENB3-256 and ENB3-512 and primarily analyze the failure cases. The obvious observation is that samples with smaller digits are relatively harder for both the models to handle. Interestingly, the formulation of UltraMNIST classification problem strictly enforces that all digits are recognized correctly, and the occurrence of even one small digit makes the classification hard. At this resolution, the largest digits are still within the scope of the receptive field of the chosen models and can be handled well. For more details, see Appendix \ref{sec-app-results}.

We would further like to reiterate on the scope of the UltraMNIST benchmark presented in this paper. Although we presented images of dimension $4000 \times 4000$, the motivation to design the benchmark comes from very large images. In this regard, it is certainly possible to generate a scaled version of UltraMNIST samples for even larger images\footnote{Code related to generation of UltraMNIST samples is accessible at \texttt{https://github.com/transmuteAI/ultramnist}}. Moreover, it is possible to increase the complexity of the samples with increased number of MNIST digits as well as more variations on the background, and this can be chosen based on the requirements of the problem. One important thing to note is that UltraMNIST is based on MNIST digits, which implies that the discriminative features of the models built on this dataset are not expected to be powerful enough to use for transfer learning on natural images. This can be looked as a limitation, however, it is beyond the scope of this work to improve on this aspect.

\textbf{Conclusion. }We present here our conclusions related to building the UltraMNIST benchmark. As stated earlier, UltraMNIST borrows inspiration from real-world problems, for example the domains of nanoscopy and microscopy, and is meant to motivate the development of CNN training pipelines that can handle very large images. With the use of multiple MNIST digits and added complexities in the background, we hope to have sufficiently simulated the challenges posed by the real cases. We have also presented multiple competitive baselines and demonstrated that the popular CNN backbones and the conventional training strategies are limited in terms of performance on this dataset. Lastly, we hope that the scientific community find this benchmark useful and contributes novel methods that can achieve higher performance scores on our dataset while not demanding too much in terms of GPU memory requirements.

\section{Acknowledgements}
Deepak Gupta would like to thank Texmin Foundation for the financial support under Project No. PSF-IH/1Y-026. Further, authors are grateful to NVIDIA for sponsoring the prizes for UltraMNIST Classification challenge hosted on the Kaggle platform. Special thanks to Igor Ivanov and the other participants who developed interesting solutions as part of the Kaggle challenge. 

{
\small
\bibliography{arxiv.bib}
}

\appendix

\section{UltraMNIST dataset}

\textbf{Terms of use, privacy and license. }This dataset is released under the terms of \texttt{Creative Commons Attribution-Share Alike 3.0} license. It implies that any copies or modifications of this work should be released under the same of similar terms, and not more restrictive terms. 

\textbf{Data maintenance. }UltraMNIST can be downloaded from the official page hosted at \texttt{https://www.kaggle.com/competitions/ultra-mnist/data}. Information related to alternative download links will also be made available on the official github page at \texttt{https://github.com/transmuteAI/ultramnist}. This page will also provide any future updates related to the dataset.

\textbf{Benchmark and code. }Code related to all benchmark experiments can be obtained from our repository at \texttt{https://github.com/transmuteAI/ultramnist}. 



\section{Experiments: Additional details}

\subsection{Implementation details}

\textbf{Data split. }For all the experiments (unless stated otherwise), we use a split of 90\% and 10\% for the train and validation sets, respectively. This validation set is used to perform early stopping of the training process and well as select model to test on the test set. 

\textbf{Training details. }All our baseline networks are trained using Adam optimizer \cite{Kingma2015AdamAM} and exponential learning rate scheduler. Further, we don't apply any data pre-processing and augmentations apart from resizing and normalization. For the case of budget-constrained UltraMNIST experiments, the batch-size of networks are scaled such that they would fill the maximum GPU memory. Further, we did random hyper-parameter search to get the best set of learning rate and gamma values for the optimizer and the scheduler, respectively. The best sets of hyper-parameters for 11 GB and 24 GB are reported in Table \ref{tab:hp11GB} and Table \ref{tab:hp24GB}, respectively.

\begin{table}[h]
\centering
\begin{tabular}{lrrrr}
\toprule
\textbf{Model}  & \textbf{Image Size} & \multicolumn{1}{l}{\textbf{Learning Rate}} & \multicolumn{1}{l}{\textbf{Gamma}} & \multicolumn{1}{l}{\textbf{Batch-size}} \\ \cmidrule(l{2pt}r{2pt}){1-5} 
               & 256  & $1\times10^{-3}$                                      & 0.8                                & 76                                      \\
EfficientNetB0 & 512  & $1\times10^{-3}$                                      & 0.8                                & 20                                      \\
               & 1024 & $1\times10^{-4}$                                      & 0.94                               & 4                                       \\\midrule

               & 256  & $1\times10^{-3}$                                      & 0.8                                & 40                                      \\
EfficientNetB3 & 512  & $5\times10^{-4}$                                      & 0.97                               & 12                                      \\
               & 1024 & $5\times10^{-5}$                                      & 0.97                               & 2                                       \\\midrule

               & 256  & $2\times10^{-3}$                                      & 1                                  & 68                                      \\
ResNet-50      & 512  & $1\times10^{-3}$                                      & 0.94                               & 18                                      \\
               & 1024 & $1\times10^{-3}$                                      & 0.97                               & 3                                       \\
\bottomrule
\end{tabular}
\vspace{0.5em}
\caption{Hyper-parameters used for 11 GB benchmark.}
\label{tab:hp11GB}
\end{table}

\begin{table}[h]
\centering
\begin{tabular}{lrrrr}
\toprule
\textbf{Model}  & \textbf{Image Size} & \multicolumn{1}{l}{\textbf{Learning Rate}} & \multicolumn{1}{l}{\textbf{Gamma}} & \multicolumn{1}{l}{\textbf{Batch-size}} \\ \cmidrule(l{2pt}r{2pt}){1-5} 
               & 256  & $4\times10^{-3}$                                      & 0.98                                & 170                                     \\
EfficientNetB0 & 512  & $2\times10^{-3}$                                      & 0.98                                & 46                                      \\
               & 1024 & $2\times10^{-3}$                                      & 0.98                                & 10                                      \\\midrule
               
               & 256  & $2\times10^{-3}$                                      & 0.98                                & 92                                      \\
EfficientNetB3 & 512  & $8\times10^{-4}$                                      & 0.98                                & 32                                      \\
               & 1024 & $2\times10^{-3}$                                      & 0.98                                & 6                                       \\ \midrule
               
               & 256  & $2\times10^{-3}$                                      & 1                                   & 140                                     \\
ResNet-50      & 512  & $2\times10^{-3}$                                      & 1                                   & 48                                      \\
               & 1024 & $3\times10^{-4}$                                      & 0.98                                & 10                                      \\
\bottomrule
\end{tabular}
\vspace{0.5em}
\caption{Hyper-parameters used for 24 GB benchmark.}
\label{tab:hp24GB}
\end{table}

\textbf{Hardware configuration. }All experiments (except MBNet-demir and EffNetTPU-vexcoz) were performed on 2080Ti (11GB) and 3090Ti (24GB) GPUs. Both the machines had 16 GB RAM and 4 core processor. EffNetTPU-vexcoz method used a Kaggle publicly available TPUv3 instance where as MBNet-demir used Kaggle publicly available P100 GPU (16 GB) instance. Both these instances have 12 GB RAM and 2 core processor.

\section{Additional Results}
\label{sec-app-results}

\textbf{Budget-constraint of 24 GB GPU memory. }We report here additional results for the case where a constraint of 24 GB is set on the maximum GPU memory to be used during the training and inference processes by the model. Results related to this experiment are reported in Table \ref{table-budumnist}. Note that the use of higher GPU memory in our case implies larger batch size, and this may not necessarily increase the performance of the model. Thus, we report two different scores for the 24 GB GPU memory budget. First is Acc$^*$@24 where the full GPU memory is utilized through scaling the batch size. Another is Acc@24 which denotes the best performance score with a 24 GB or less of GPU memory. 

Looking at the results, we see cases where utilizing the full GPU memory does not lead to increased performance. On that contrary, we see that for RN-256 and ENB3-512, performance drops. For RN-256, we have observed that the discriminative power of the model is not good enough to capture the fine details of the images. Due to extreme downscaling to $256\times 256$, the small digits are hard to identify, and this leads to lower performance of RN-256. For the case of larger batch size, we believe that with further tuning of the learning rate and its decay, the performance gap between Acc$^*$@24 GB and Acc@24 GB could be eliminated. In terms of the overall best for 24 GB budget, we have MBNet-demir which uses MobileNet-V3 as well as only utilizes a maximum memory of 16 GB. 

\begin{table}[]
    \centering
    \begin{tabular}{l l c c c c}
         \toprule
         \textbf{Method} & \textbf{Backbone Model} & \textbf{Input size} & \textbf{Acc$^*$@24 GB} \% & \textbf{Acc.@24 GB} \% \\
        \midrule
         RN-256 & ResNet-50 & 256$\times$256 & 5.2 & 8.9\\
         RN-512 & ResNet-50 & 512$\times$512 & 24.8 & 24.8\\
         ENB0-256 & EfficientNet-B0 & 256$\times$256 & 20.4 & 20.4 \\
         ENB0-512 & EfficientNet-B0 & 512$\times$512 & 49.9 & 49.9\\
         ENB0-1024 & EfficientNet-B0 & 1024$\times$1024 & 52.3 & 52.3\\
         ENB3-256 & EfficientNet-B3 & 256$\times$256 & 30.3 & 30.3\\
         ENB3-512 & EfficientNet-B3 & 512$\times$512  & 67.5 & 69.1\\
         MBNet-demir & Mobilenet-V3 & 1280$\times$1280 & - & \textbf{86.2} \\
         \bottomrule
    \end{tabular}
    \vspace{0.5em}
    \caption{Performance scores for various baselines on UltraMNIST benchmark. Here \mbox{Acc$^*$@24 GB \%} denotes the accuracy score in \% obtained for full utilization of 24 GB GPU memory. Further, Acc@24 GB \% denotes the best accuracy score for the GPU utilization of 24 GB memory or even lesser (\emph{e.g.} 16 GB, 11 GB, etc).}
    \label{table-budumnist}
\end{table}

\textbf{Analysis of failure cases. }To get a better understanding of when and why the model tends to mostly fail, we analyzed some of the failure cases for the results obtained on the validation set using ENB0-256 and ENB0-512 models. For this purpose, we randomly sampled 20 cases for each model where the prediction of the model was incorrect. For both the cases, we observed that the model is able to identify the larger digits even under the huge clutter. However, most of the failure cases happen in presence of the small digits. For the case of $256\times256$ the occurrence of small digits make the recognition extremely hard due to which we see significant drop in performance for ENB0-256. This clearly indicates that downsampling is not the best direction to obtain maximum performance and we need to identify ways to train CNN methods which images that undergo no downsampling.

\textbf{Effect of random seed. }UltraMNIST dataset contains significant noise in each image. Moreover, the occurrence of extremely small digits together with the larger ones makes the performance of the model very sensitive to the training process. In our experience, even small differences in the training regime could significantly affect the performance. To quantify this affect, we conducted a small sensitivity analysis test to investigate how much the accuracy of the model is affected by the random seed. For this, we used the EfficientNetB0  architecture with image resolutions of $256\times 256$ and $512\times 512$. We found that ENB0-256 obtained accuracy of 18.45$\pm$1.6\% whereas ENB0-512 got 43.85$\pm$4.7\% over 3 runs. Clearly, there is a huge variance in the performance of the models for the variation in the random seed, and this has to do with the complexity of the information in the UltraMNIST samples. The high standard deviation in these experiments shows that we need better high-resolution image training pipelines and the conventional pipelines are not well suited for such tasks, at least from the stability perspective. We intend to conduct a more detailed research on this aspect as a part of our future work. 

\textbf{Unconstrained UltraMNIST classification. }The unconstrained variant of our UltraMNIST competition refers to the pool where there is no restrictions are imposed. This implies that the candidates are free to use the MNIST dataset for model training, transform UltraMNIST into an object detection problem to identify each digit separately and then to build a model that learns to compute the summation. The use of MNIST digits allows building a replica of UltraMNIST where the bounding box around each digit is known for the training set, and object detectors can be trained. With extreme engineering, participants were able to score accuracy scores of even as high as 99\%. Further discussion  can be found at \texttt{https://www.kaggle.com/competitions/ultra-mnist/discussion}.

Note that the focus of this paper is on the standard UltraMNIST and its budget-aware variants, and we hope to see developments along improved CNN training pipelines that can handle very large images.
\begin{figure}[h]
    \centering
    \begin{subfigure}{1.0\linewidth}
    \centering
    \fbox{\includegraphics[scale=0.022]{images/samples/0/dngoaadrfn.jpeg}}
    \fbox{\includegraphics[scale=0.022]{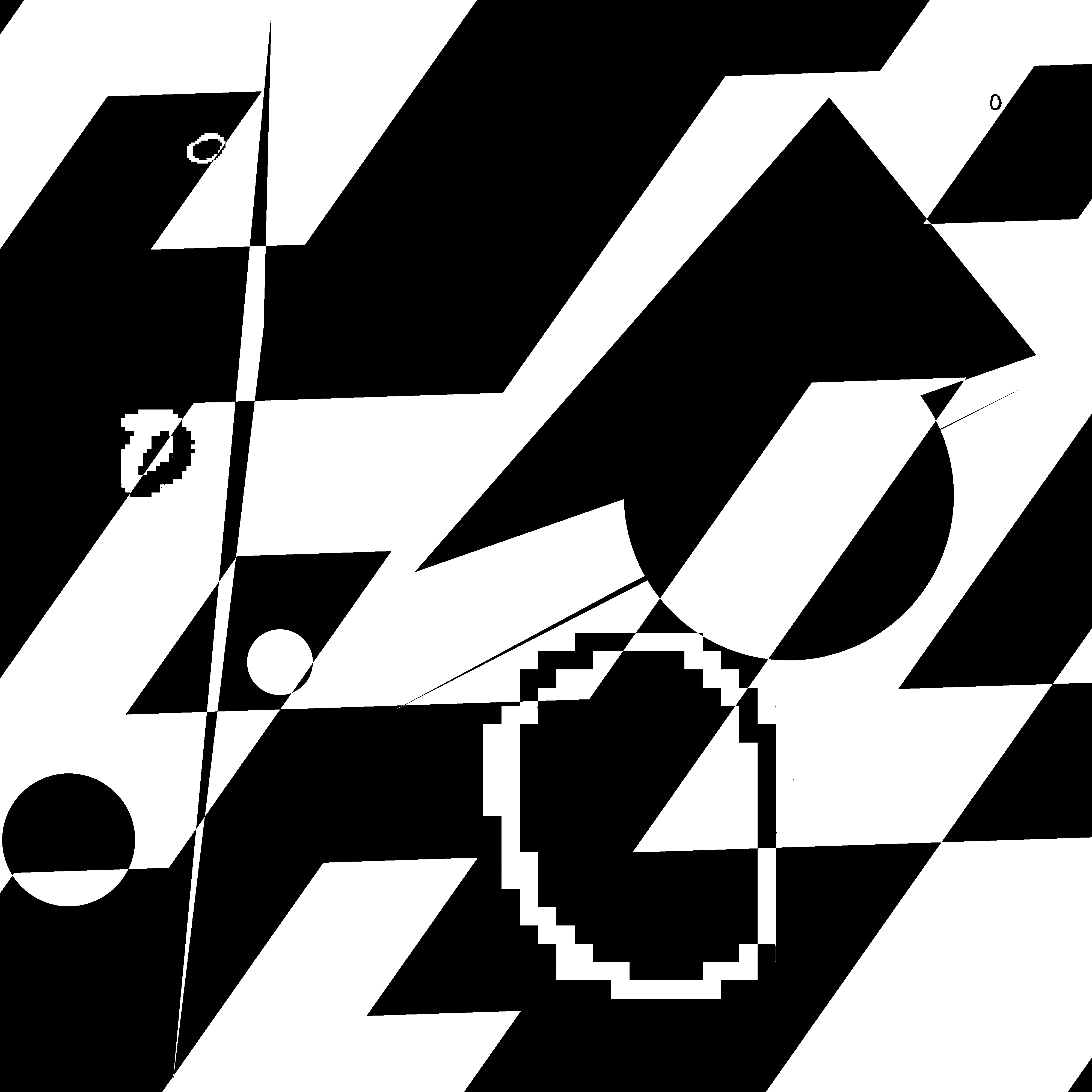}}
    \fbox{\includegraphics[scale=0.022]{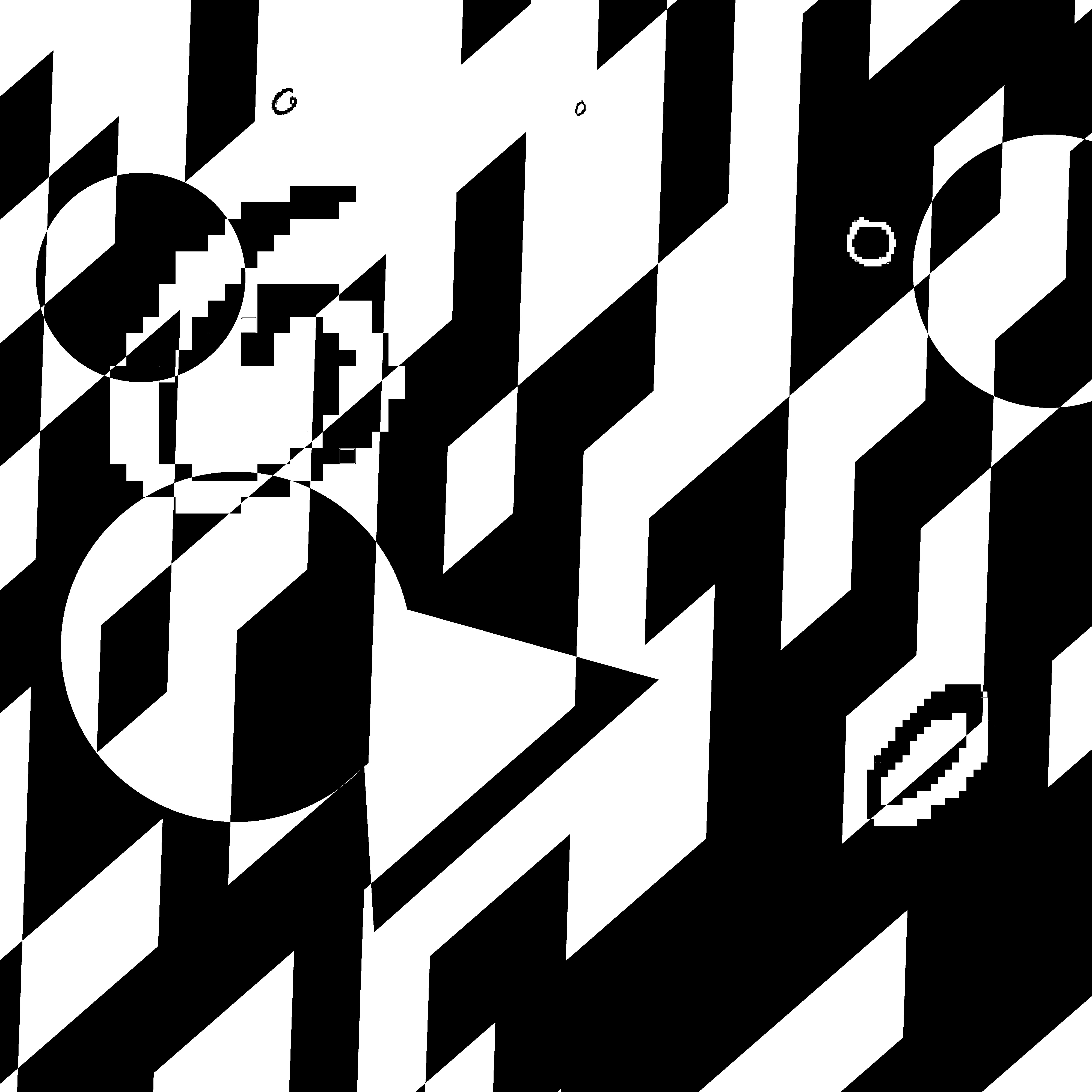}}
    \fbox{\includegraphics[scale=0.022]{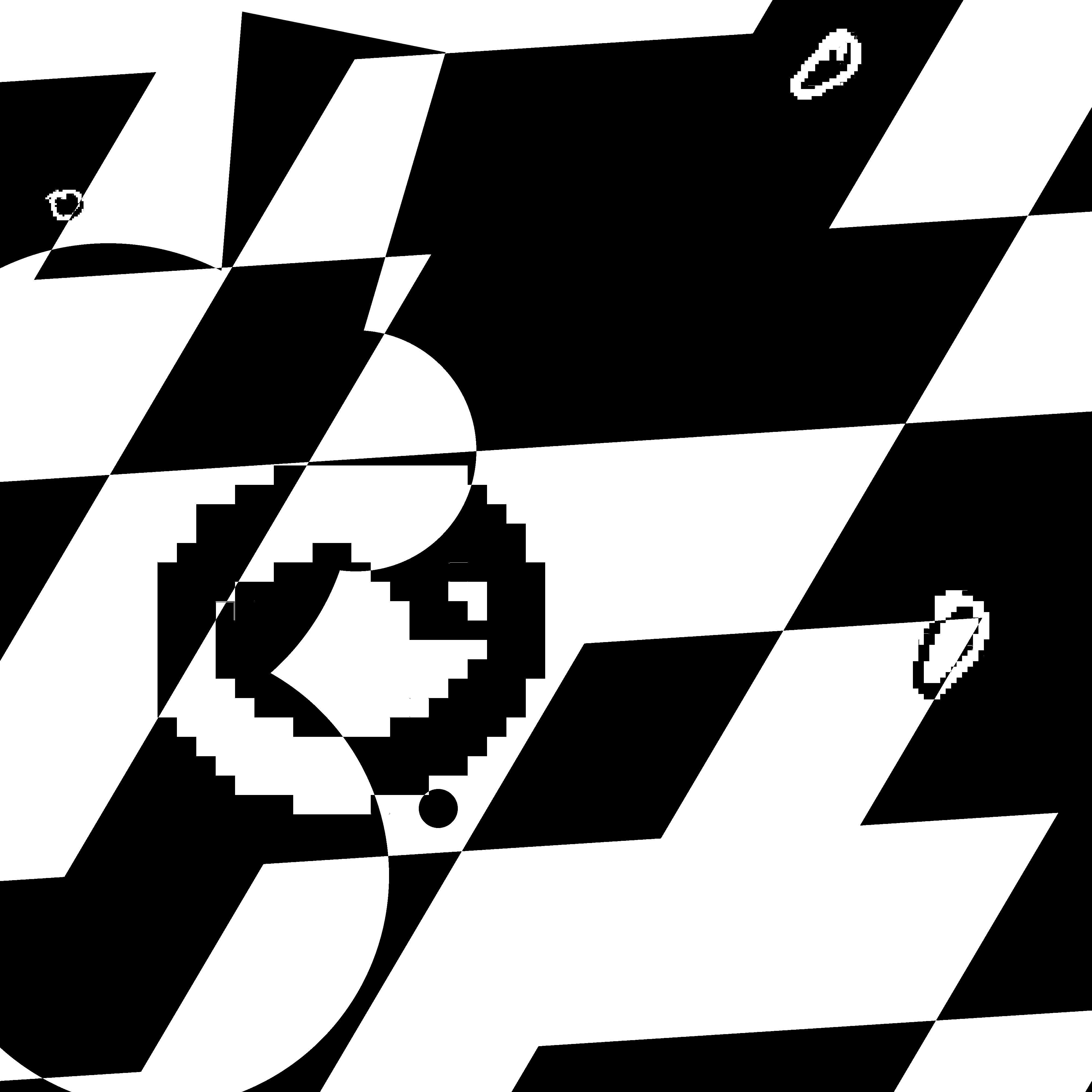}}
    \caption{Samples of Class 0}
    \end{subfigure}
    \begin{subfigure}{1.0\linewidth}
    \centering
    \fbox{\includegraphics[scale=0.022]{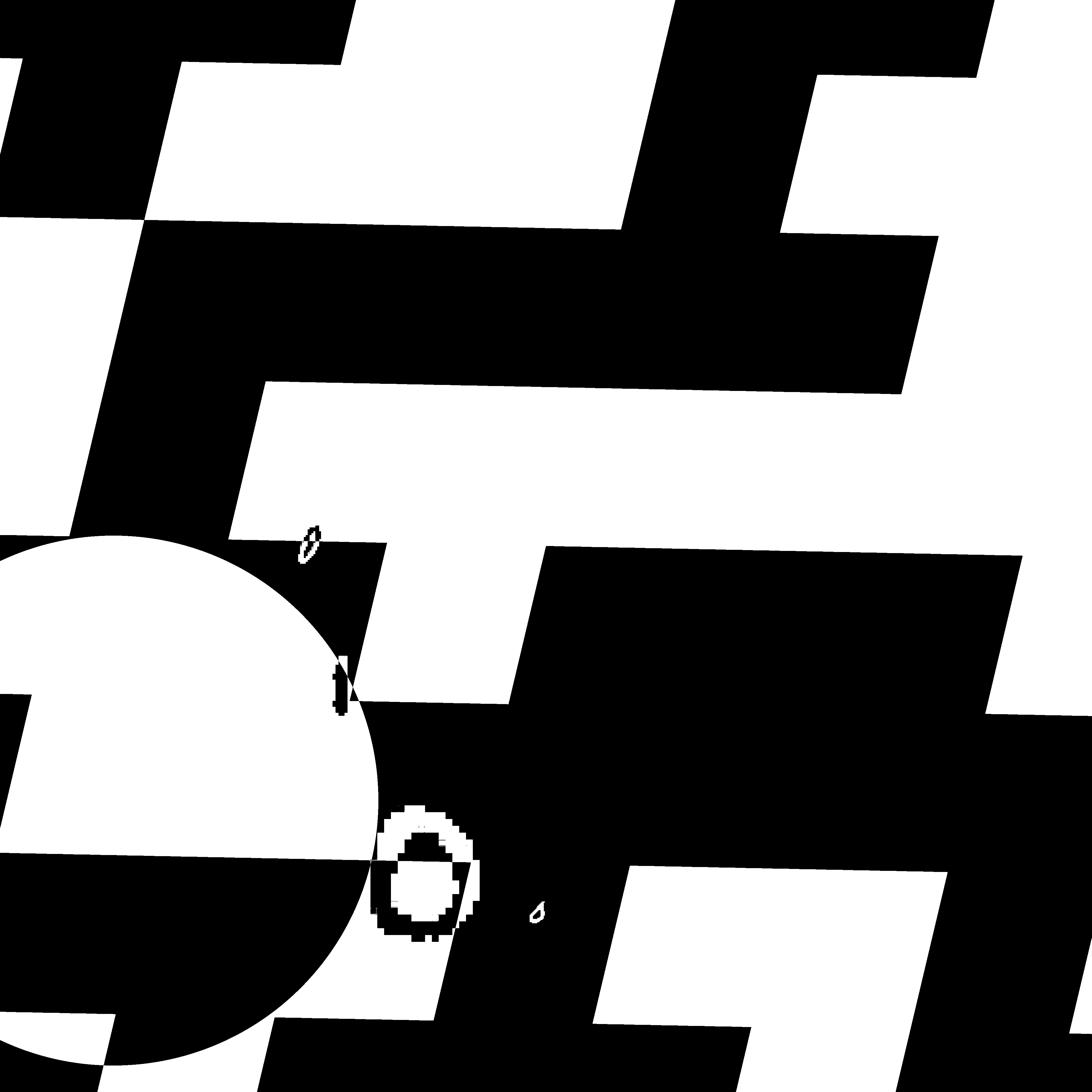}}
    \fbox{\includegraphics[scale=0.022]{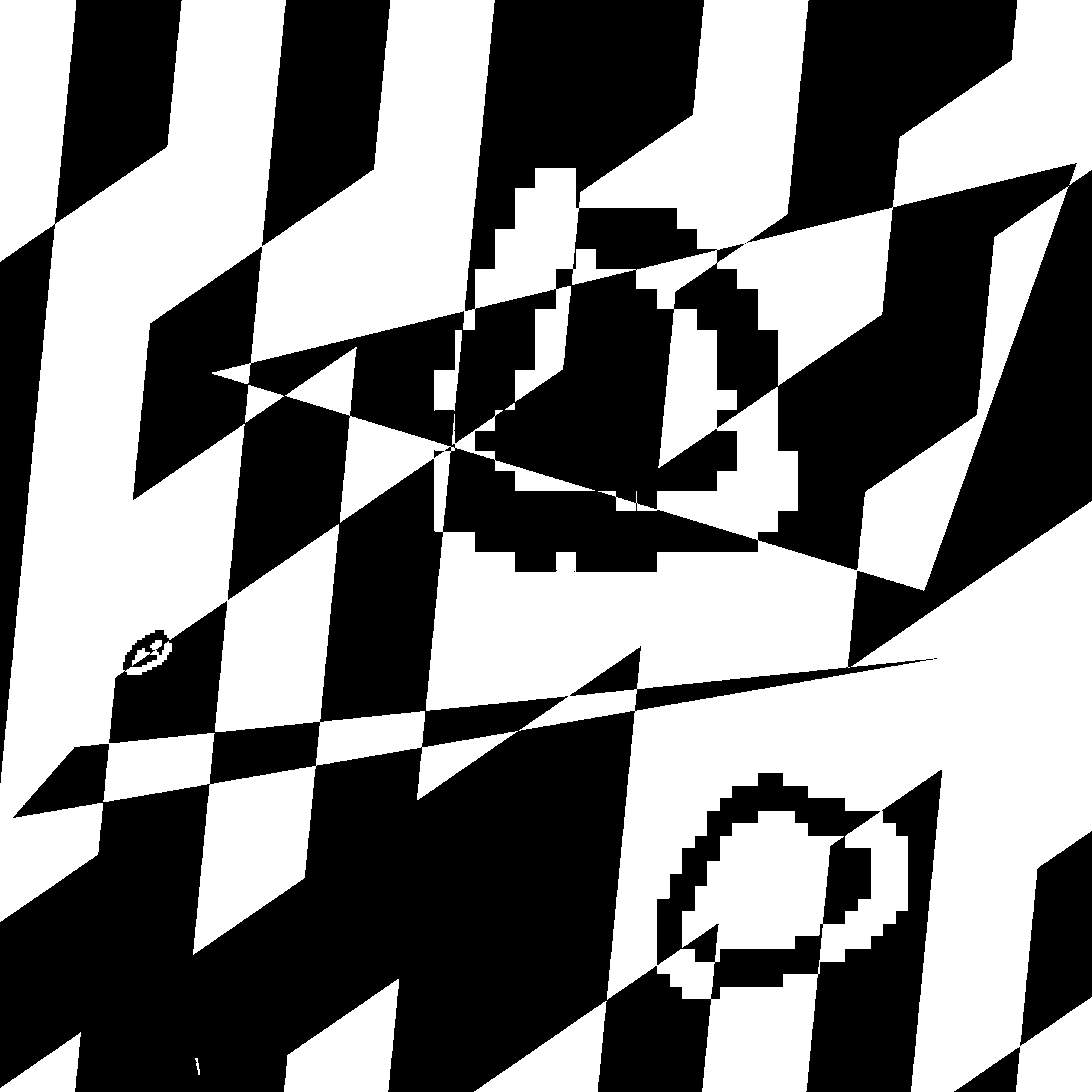}}
    \fbox{\includegraphics[scale=0.022]{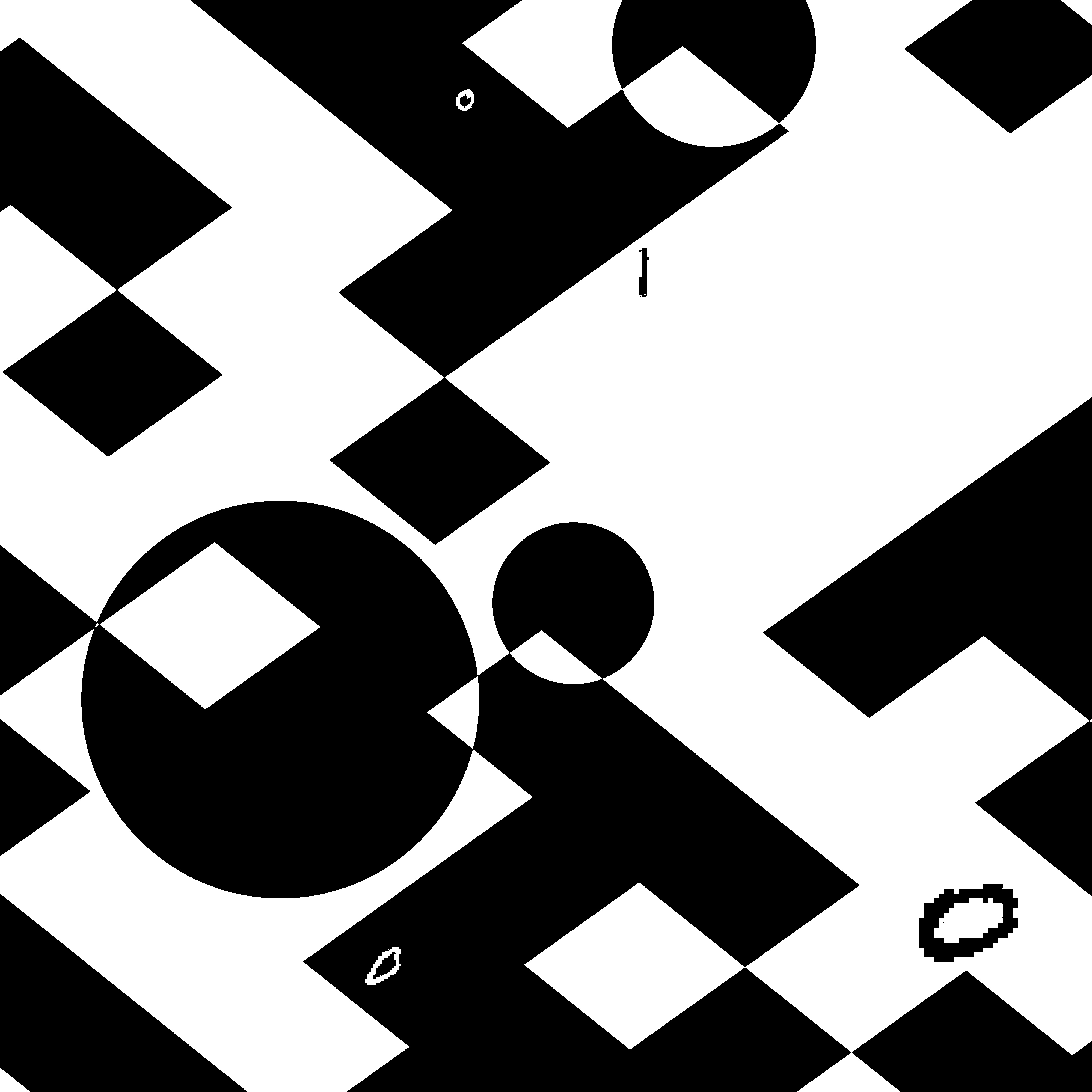}}
    \fbox{\includegraphics[scale=0.022]{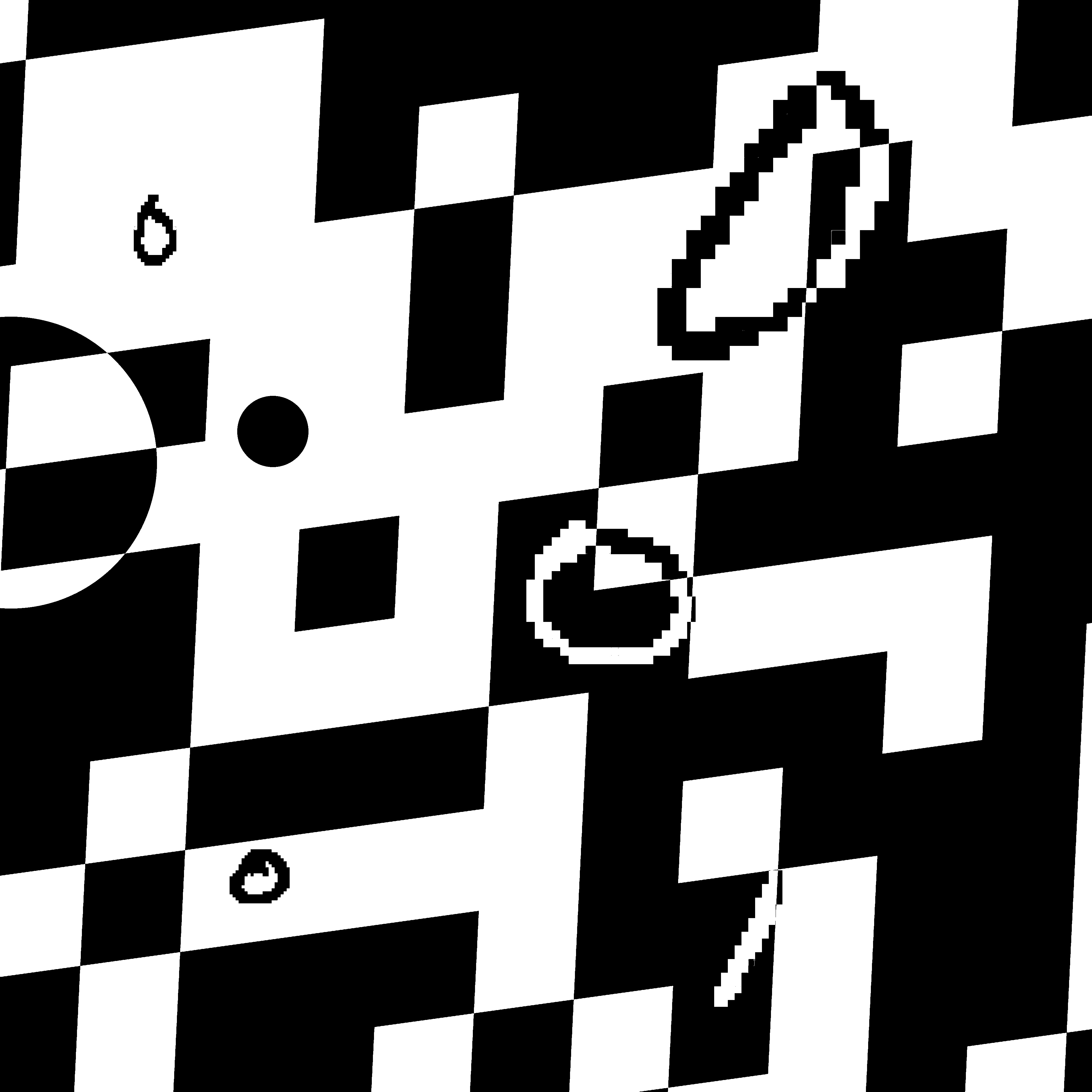}}
    \caption{Samples of Class 1}
    \end{subfigure}
    \begin{subfigure}{1.0\linewidth}
    \centering
    \fbox{\includegraphics[scale=0.022]{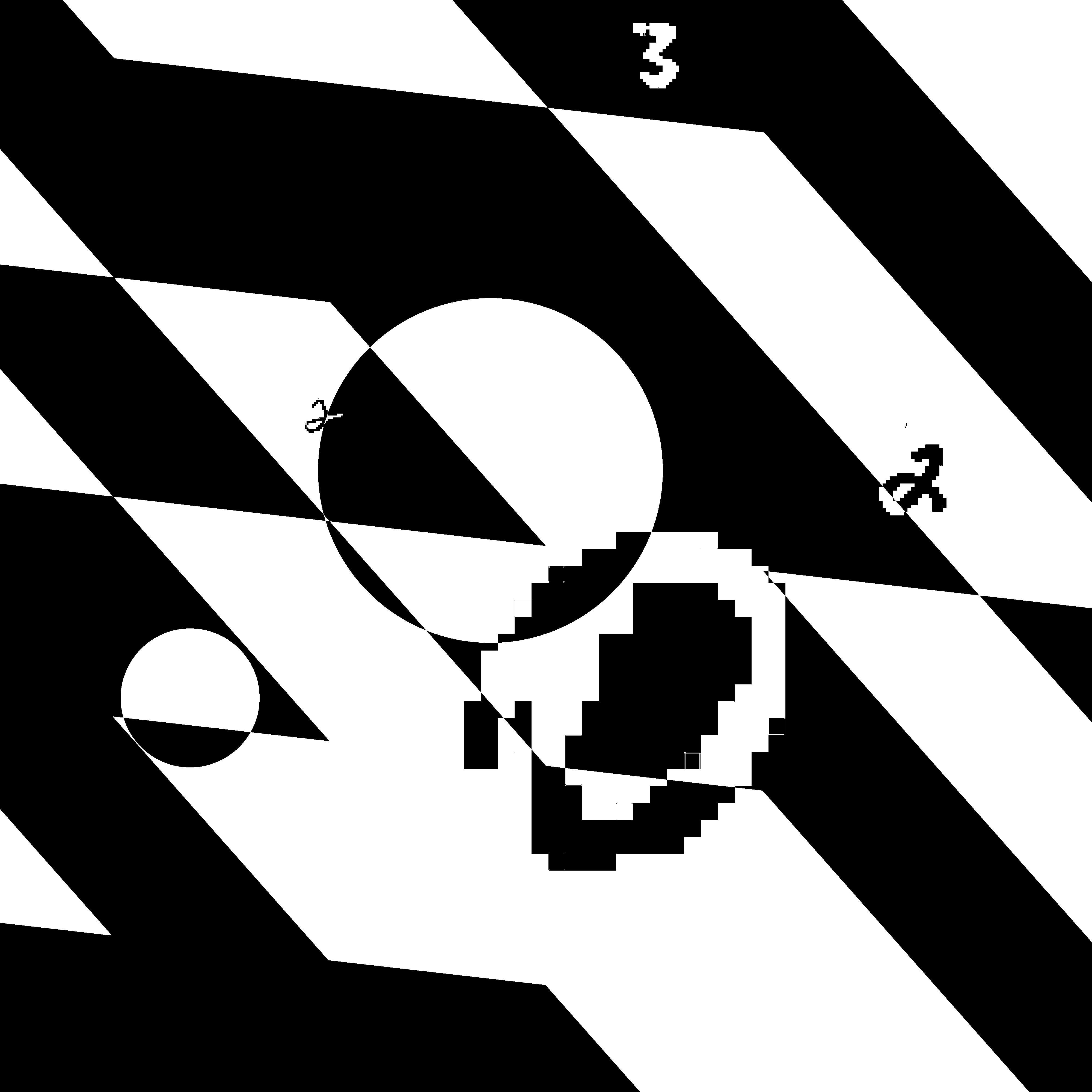}}
    \fbox{\includegraphics[scale=0.022]{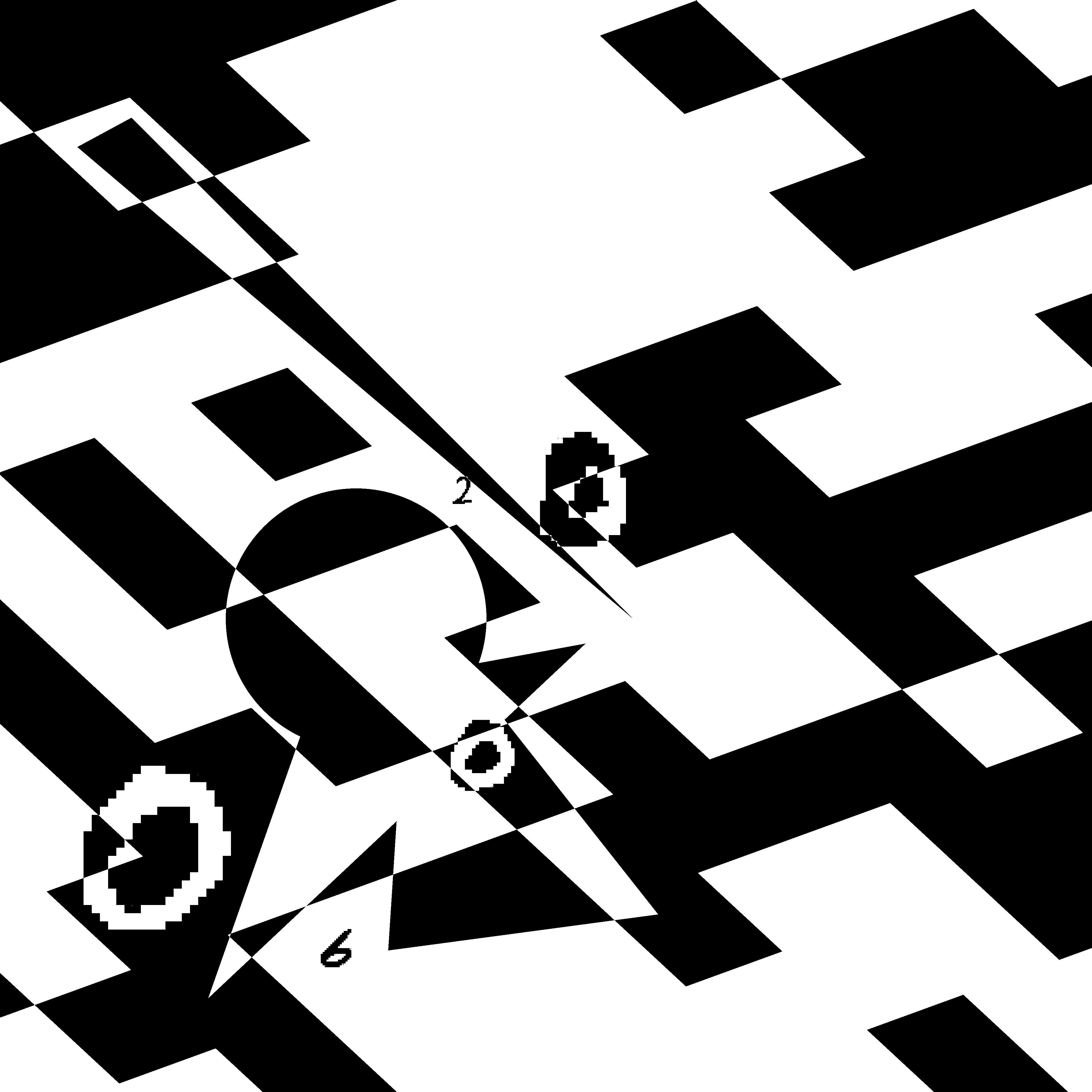}}
    \fbox{\includegraphics[scale=0.022]{images/samples/8/befsxjymof.jpeg}}
    \fbox{\includegraphics[scale=0.022]{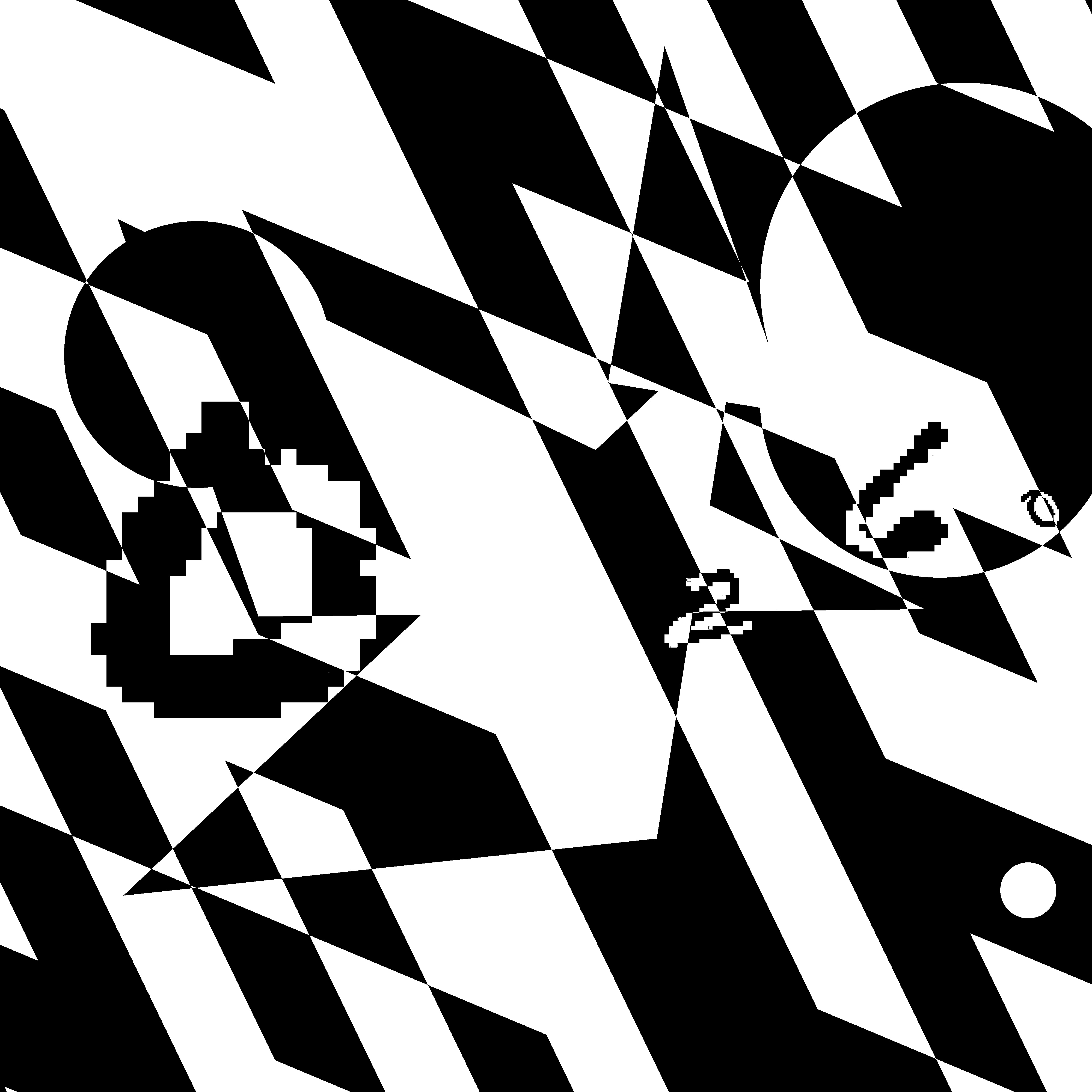}}
    \caption{Samples of Class 8}
    \end{subfigure}
    \begin{subfigure}{1.0\linewidth}
    \centering
    \fbox{\includegraphics[scale=0.022]{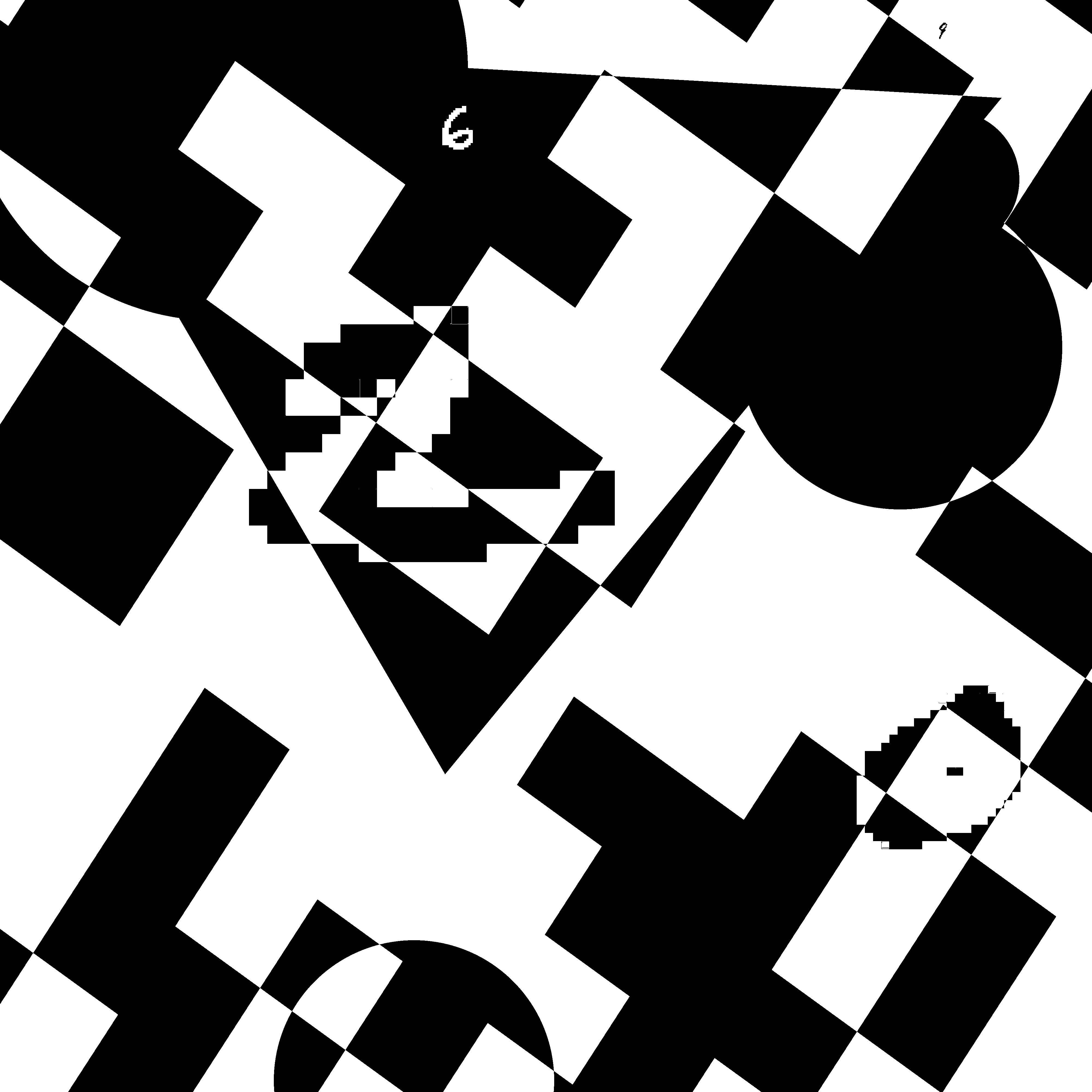}}
    \fbox{\includegraphics[scale=0.022]{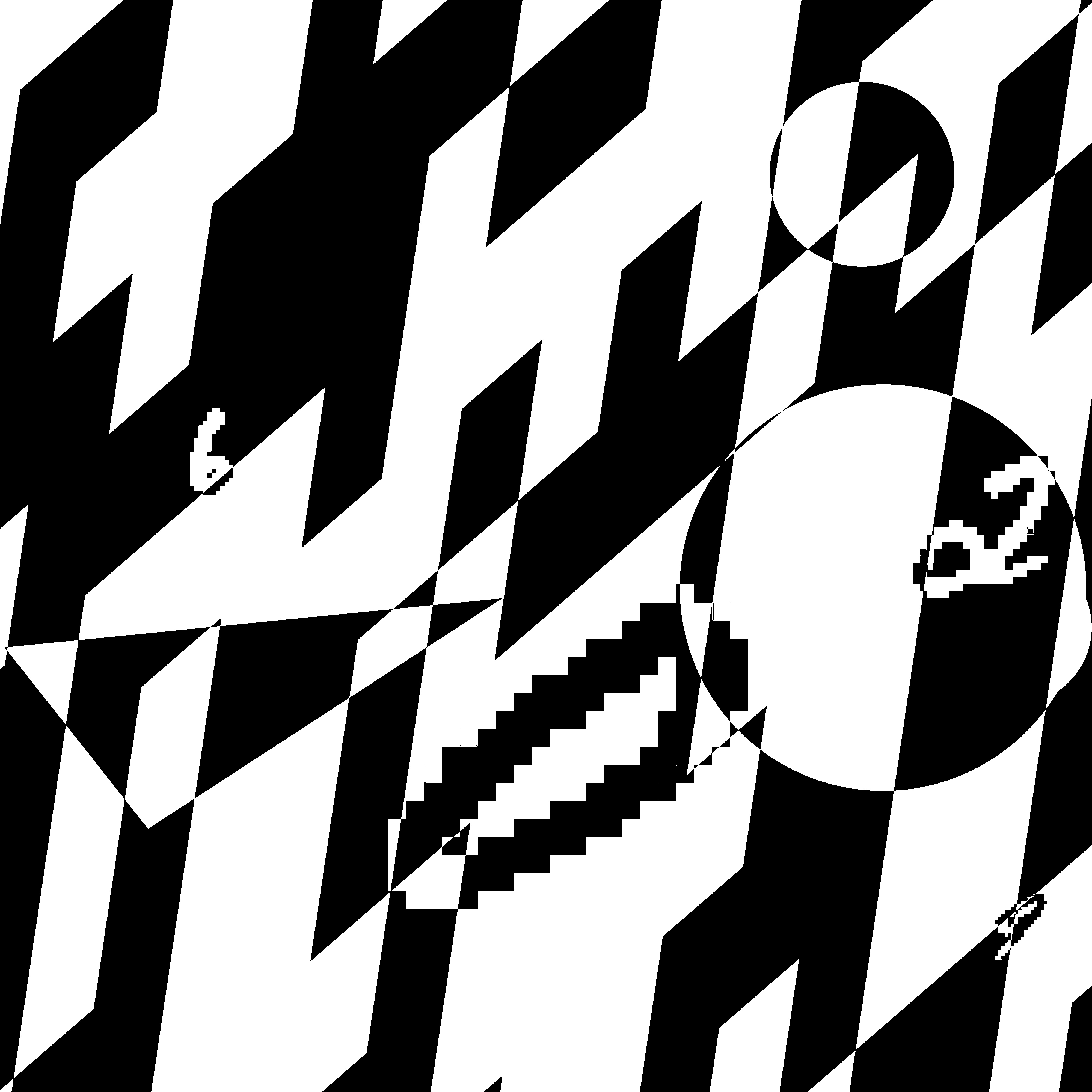}}
    \fbox{\includegraphics[scale=0.022]{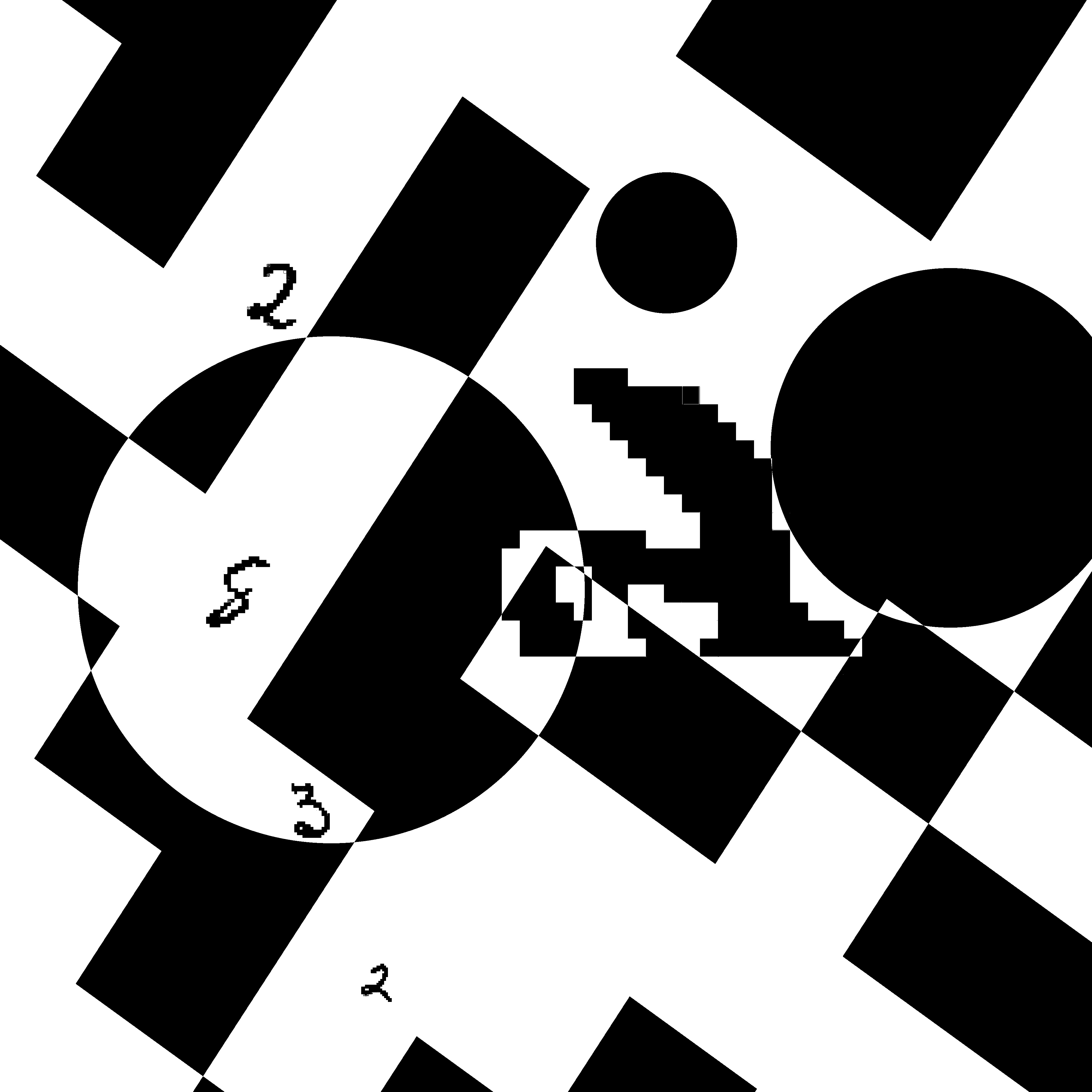}}
    \fbox{\includegraphics[scale=0.022]{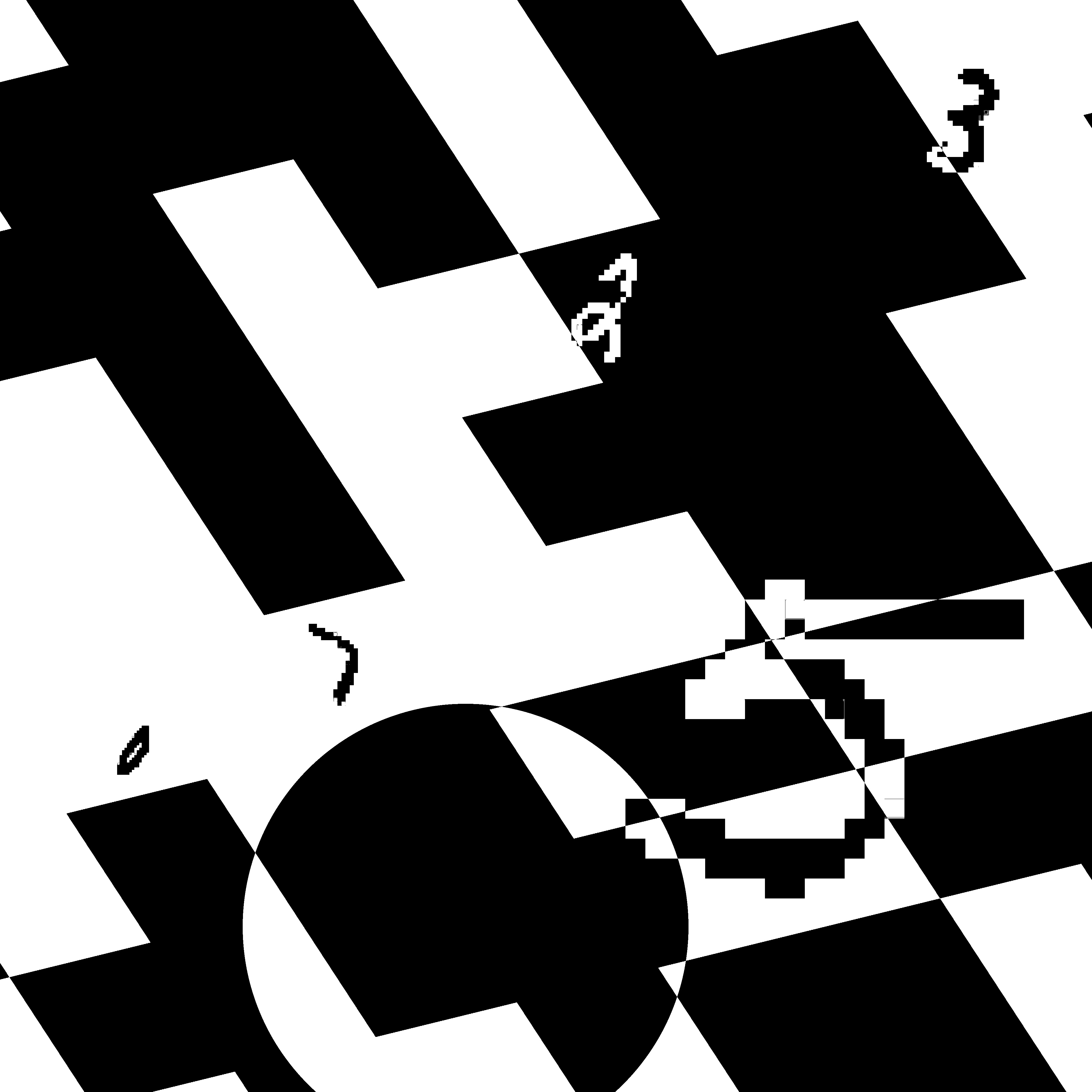}}
    \caption{Samples of Class 17}
    \end{subfigure}
    \begin{subfigure}{1.0\linewidth}
    \centering
    \fbox{\includegraphics[scale=0.022]{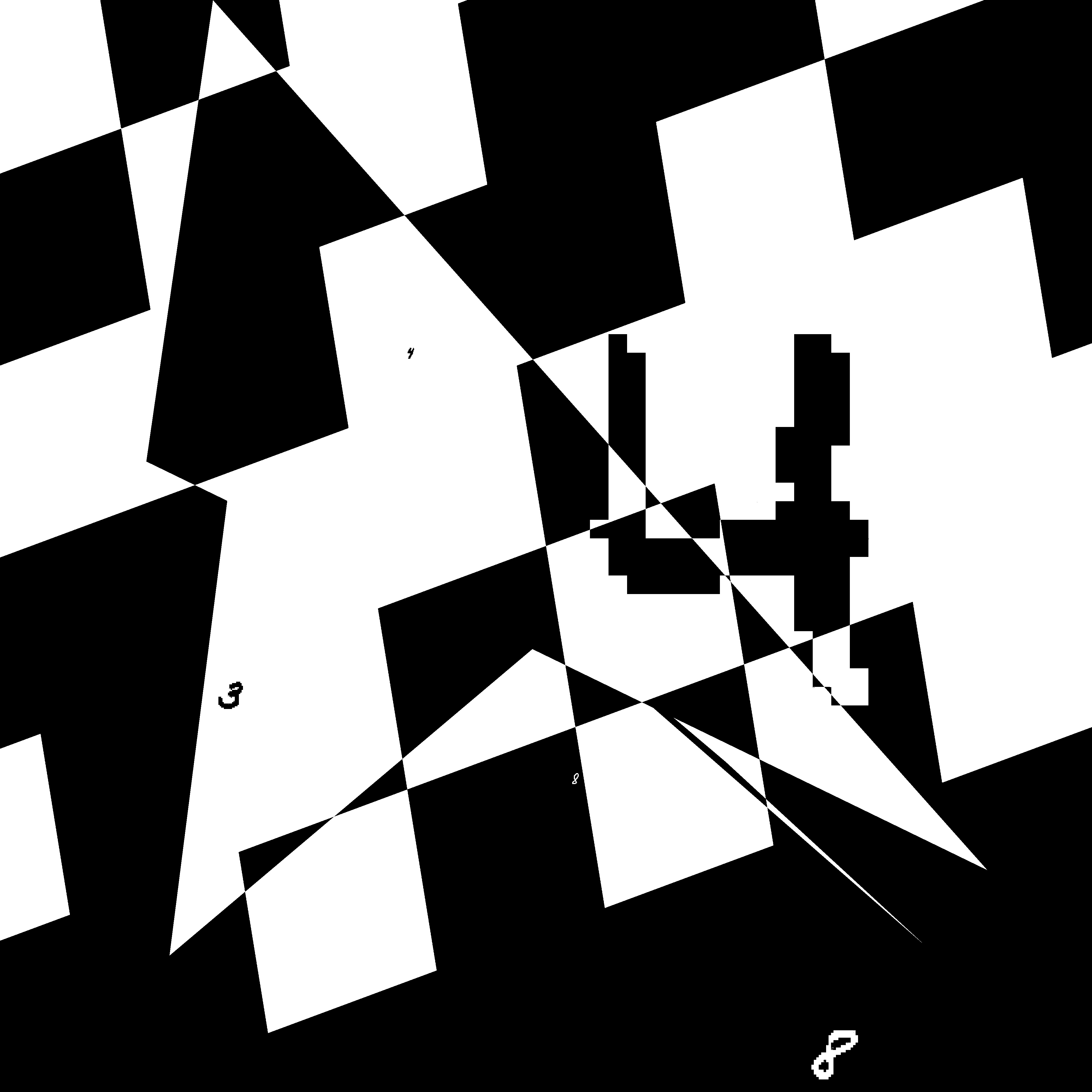}}
    \fbox{\includegraphics[scale=0.022]{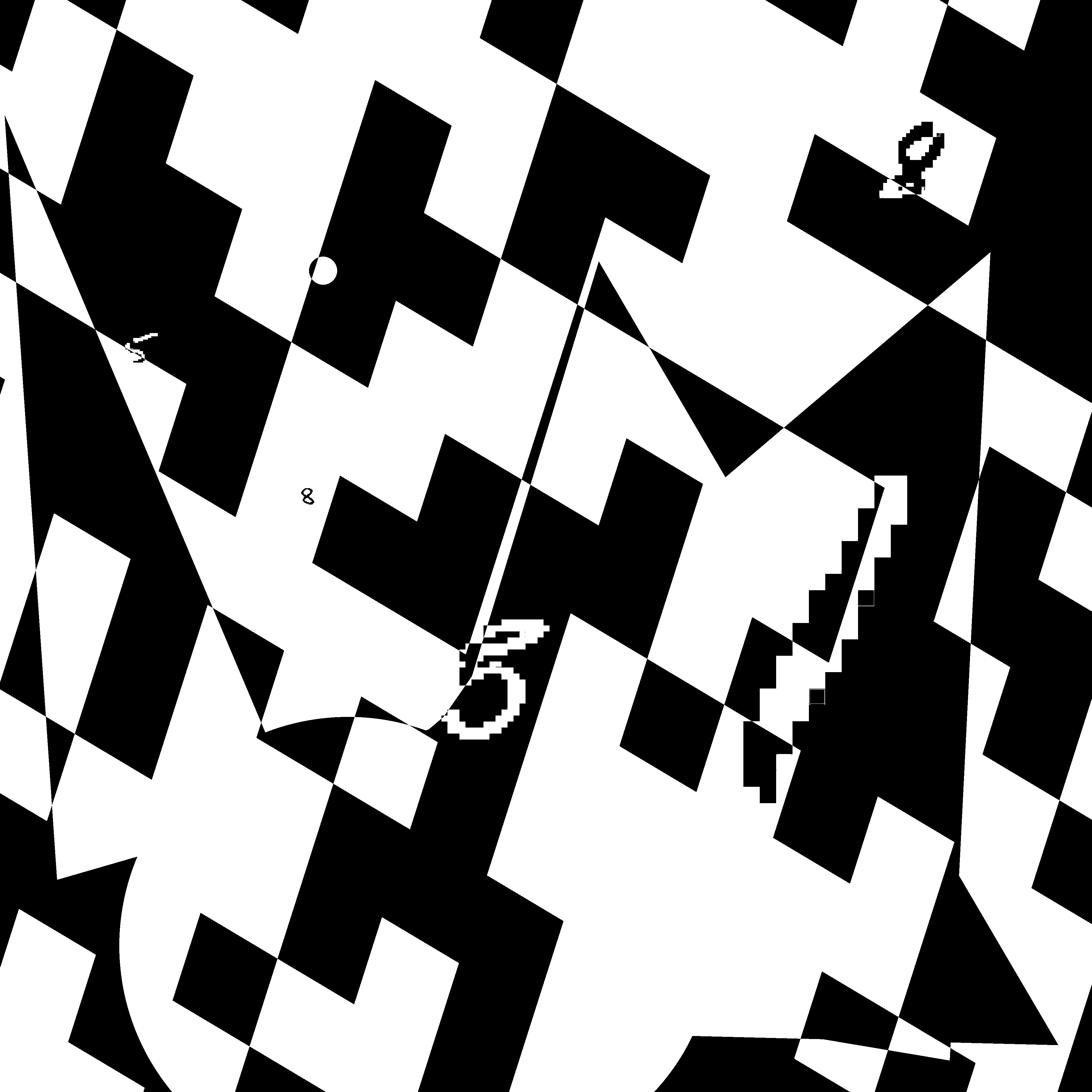}}
    \fbox{\includegraphics[scale=0.022]{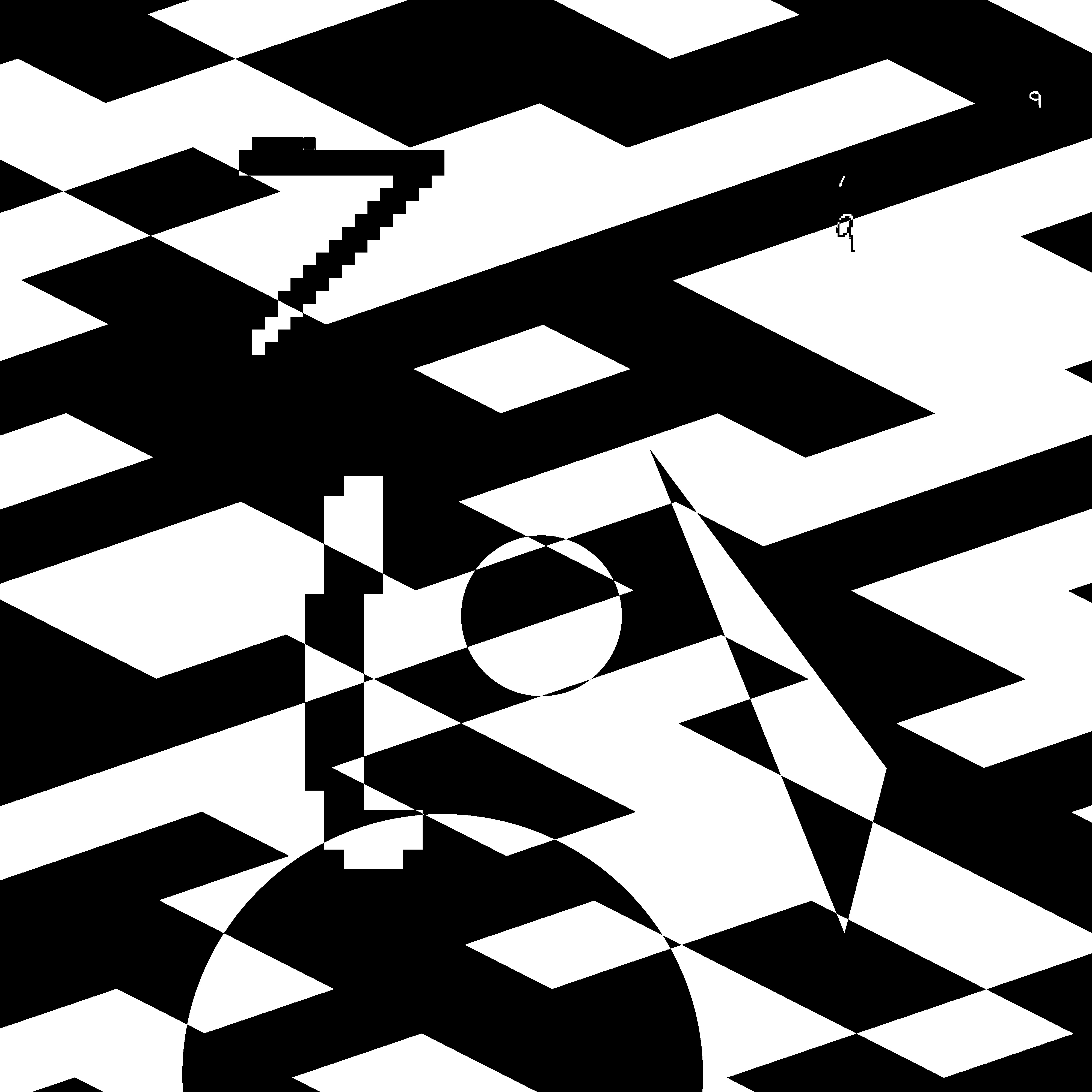}}
    \fbox{\includegraphics[scale=0.022]{images/samples/27/fmkdekzxjj.jpeg}}
    \caption{Samples of Class 27}
    \end{subfigure}
    \caption{Examples from various classes of UltraMNNIST dataset.}
    \label{fig:app-samples}
\end{figure}

\textbf{Additional visualizations. }For the visualization of additional samples from different classes of the UltraMNIST dataset, please refer Figure \ref{fig:app-samples}.

\end{document}